\documentclass[runningheads]{llncs}

 \usepackage{eccv}

\newcommand{\dsname}{O3-D}

\usepackage{eccvabbrv}

\usepackage{graphicx}
\usepackage{booktabs}
\usepackage{enumitem}

\usepackage{pifont}
\newcommand{\cmark}{\ding{51}}
\newcommand{\xmark}{\ding{55}}

\usepackage{eso-pic}
\usepackage{xcolor} 

\newcommand{\preprintnotice}{%
  \AddToShipoutPictureBG*{
    \AtPageUpperLeft{%
      \put(0,\LenToUnit{-1.5cm}){
        \parbox[t]{\paperwidth}{%
          \centering
          \normalfont\sffamily\fontsize{9}{11}\selectfont\color{gray}
          Accepted to the 19th European Conference on Computer Vision -- ECCV 2026
        }%
      }%
    }%
  }%
}

\usepackage[accsupp]{axessibility}  

\usepackage{hyperref}
\hypersetup{
    bookmarksdepth=2  
}
\usepackage{bookmark} 

\usepackage{orcidlink}

\begin{document}

\title{Disentangling Pictorial Cue Understanding from Language Bias in VLMs via Depth Ordering Task}

\titlerunning{Disentangling Pictorial Cue Understanding from Language Bias}

\author{Yiqian Liu \inst{1}\orcidlink{0009-0001-3575-3454} \and
Iuliia Kotseruba\inst{2}\orcidlink{0000-0002-5490-9805} \and
John K. Tsotsos\inst{1}\orcidlink{0000-0002-8621-9147}}

\authorrunning{Y.~Liu et al.}

\institute{York University, Toronto, Canada
\email{\{yql,tsotsos\}@yorku.ca}\\
\and
University of Guelph, Guelph, Canada\\
\email{kotseruba@uoguelph.ca}}

\maketitle
\preprintnotice

\begin{abstract}
In this paper, we study depth perception of vision-language models (VLMs) to isolate the effects of pictorial depth cues and disentangle vision and language influences on model performance. To this end, we combine depth-ordering and odd-one-out psychophysical tasks: the VLMs are presented with images where one object is at different depth relative to other, otherwise identical, objects, and must determine whether the odd-one-out target is closer or farther to the observer. To create stimuli, we generate 2D views from simulated and real 3D scenes while controlling the presence of individual pictorial depth cues, enabling a fine-grained analysis of cue-level contributions. Language effects are examined by varying referring expression clarity. We also introduce a novel metric to quantify vision-vs-language sensitivities. Applying this methodology, we create the Odd-One-Out Depth (O3-D) dataset with 37K real and synthetic images and 147K image-question pairs. Evaluation of 12 open-source and commercial models on O3-D shows under-utilization of depth cues and depth-ordering accuracies between 47\% and 56\%, with no model above chance level. At the same time, our metric reveals strong linguistic bias in the answers. Neither chain-of-thought (CoT) nor in-context learning (ICL) significantly improves performance, suggesting that static image data alone may be insufficient for depth understanding. All code, the image generation pipeline, and the O3-D dataset are publicly released at \url{https://github.com/lyiqian/o3-d}. 
\end{abstract}

\section{Introduction}
\label{sec:intro}
Extracting rich spatial structure from images is one of the fundamental computer vision problems. Historically, it has been approached from multiple angles, such as monocular depth estimation \cite{Arampatzakis_2023_TPAMI}, object detection \cite{zou2023object}, segmentation \cite{minaee2021image}, and 3D scene reconstruction \cite{samavati2023deep}. The most recent generation of vision-language models (VLMs) aims to perform scene understanding as a single system. 

Despite being trained only on static images, VLMs demonstrate a range of depth-related abilities. For example, past benchmarks \cite{chowPhysBenchBenchmarkingEnhancing2025, fuBLINKMultimodalLarge2025, azadUnderstandingDepthHeight2025, chenSpatialVLMEndowingVisionLanguage2024, chengSpatialRGPTGroundedSpatial2024, tongEyesWideShut2024, liuMMBenchYourMultimodal2025, liSEEDBenchBenchmarkingMultimodal2024} showed evidence of VLMs understanding size, distance, and structure of objects. However, utilization of individual depth cues by these models remains understudied. Another challenge in evaluating VLMs is posed by their inherent bimodal nature. As most VLM evaluation protocols are based on Visual Question Answering (VQA), both vision and language modalities interact, making it difficult to disentangle their individual effects on the overall performance.

\begin{figure}[tb]
  \centering
  \includegraphics[width=\textwidth]{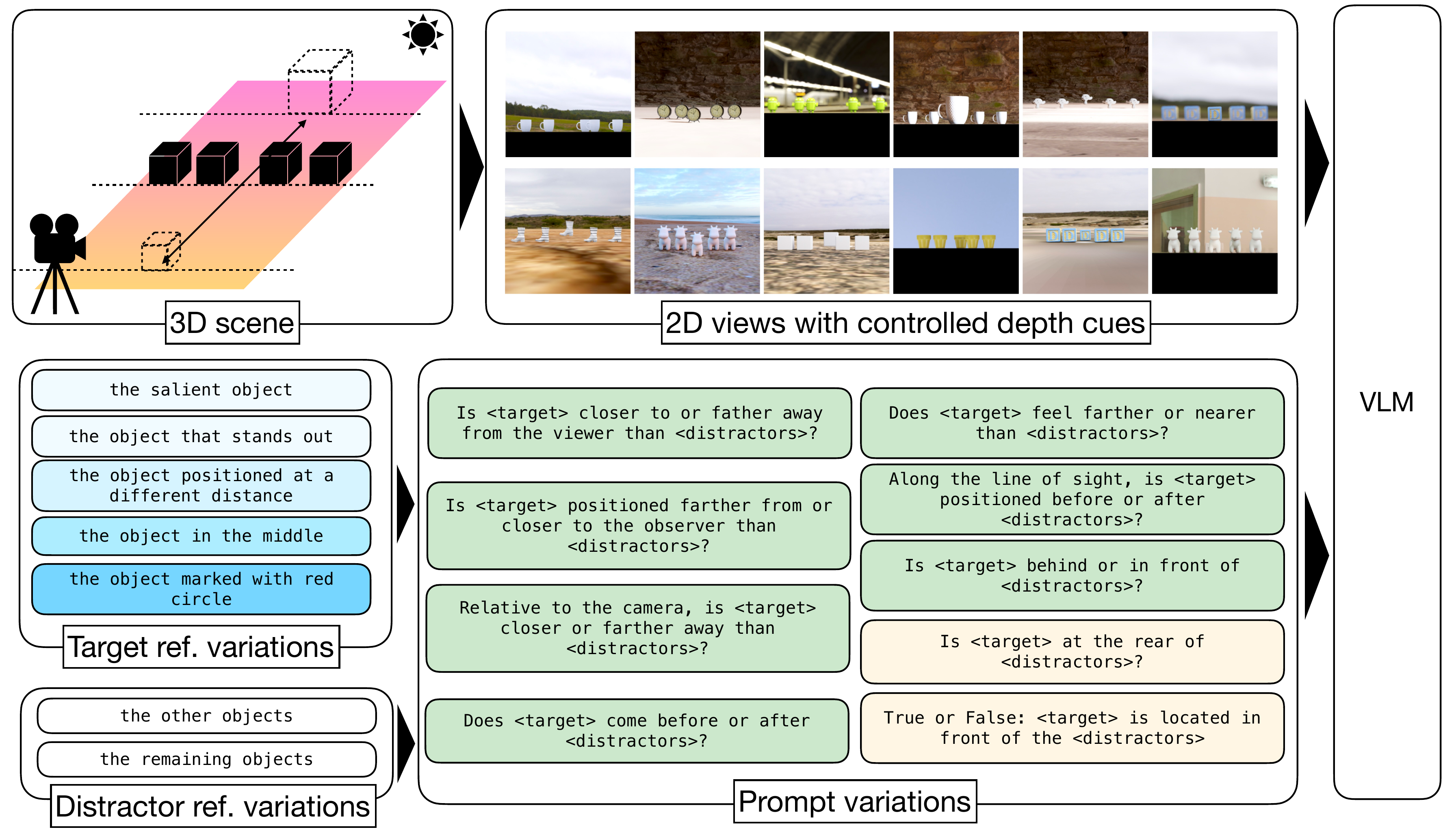}

   \caption{\dsname\ probes VLM depth and language understanding. We start by constructing synthetic and real \texttt{3D scenes} with diverse backgrounds and objects. Each scene contains 5 objects of the same class, one of which (the target) is of different size and placed at a different depth plane. We then generate a number of \texttt{2D views with one or two depth cues} by controlling the camera, light position, \etc. For each image, we create \texttt{prompt variations}: multiple-choice (shown in green) and yes-no (yellow). Within each prompt, we vary the target referring clarity (more specific references are shown in darker shades of blue in the \texttt{Target ref. variations} box) and distractor descriptions (\texttt{Distractor ref. variations}).}
   \label{fig:fig1}
\end{figure}

To address the limitations of existing depth benchmarks, we propose the {\it Odd-One-Out Depth (\dsname)} dataset and evaluation metrics for systematic evaluation of VLMs' depth understanding capabilities while taking into account complexities arising from the language dimension (\cref{fig:fig1}). We approach the problem from a psychophysical standpoint by using a depth ordering task \cite{brennerDepthPerception2018} within an odd-one-out setup \cite{sinapovOddOneOut2010}. Specifically, we construct a series of synthetic 3D odd-one-out scenes where one object is at different depth relative to other, otherwise identical, objects (\cref{fig:base-bev-a}). We then generate a number of 2D views from these 3D scenes while controlling for 9 common pictorial depth cues \cite{reicheltDepthCuesHuman2010, kavsekInfantsSensitivityPictorial2012, wattFocusCuesAffect2005} to isolate cue-level contributions (\cref{fig:base-bev-c}). Additionally, \dsname\ includes real-world odd-one-out-depth images from two sources: captured in a custom setup and selected from \cite{kotserubaSaliencyModelsDetect2021}. Along the language dimension, we consider referring expression comprehension, which commonly deals with language ambiguity \cite{fitzgeraldLearningDistributionsLogical2013}. Referring to objects is especially challenging in visual contexts with multiple similar objects, thus we experiment with the referring clarity. In addition, we introduce a novel metric for measuring the relative sensitivities of vision and language input modalities. Lastly, we test whether common techniques such as chain-of-thought (CoT) and in-context learning (ICL) can improve VLMs' depth perception. Our main contributions are summarized as follows:

\begin{itemize}[topsep=0em] 
\itemsep0em
\item We propose a challenging dataset, \dsname\, for the odd-one-out depth ordering VQA task to evaluate VLMs depth capabilities along vision and language dimensions. For the former, we test utilization of individual pictorial depth cues. For the latter, we vary the clarity of referring expression.
\item We design a novel pipeline for constructing synthetic odd-one-out scenes with configurable objects, environments, cues, and prompt variations.
\item We formulate a novel metric for measuring the relative influences of vision and language inputs on the VLMs' overall performance.
\item Through extensive experimental validation, we demonstrate low individual pictorial cue utilization as well as consistently high language influence across 12 commercial and open-source SOTA VLMs. 
\end{itemize}

\section{Related Works}

{\bf Depth understanding of foundation vision and vision-language models.}
Many works examined monocular depth understanding of vision models \cite{danierDepthCuesEvaluatingMonocular2025, manLexicon3DProbingVisual2024a, elbananiProbing3DAwareness2024a, linsley3DPCBenchmarkVisual2025, zhanGeneralProtocolProbe2024a} and VLMs \cite{chowPhysBenchBenchmarkingEnhancing2025, fuBLINKMultimodalLarge2025, azadUnderstandingDepthHeight2025, chenSpatialVLMEndowingVisionLanguage2024, chengSpatialRGPTGroundedSpatial2024, tongEyesWideShut2024, liuMMBenchYourMultimodal2025, liSEEDBenchBenchmarkingMultimodal2024}.
Generally, the depth understanding ability was probed indirectly by testing models' perception of 
size \cite{danierDepthCuesEvaluatingMonocular2025,chowPhysBenchBenchmarkingEnhancing2025,chengSpatialRGPTGroundedSpatial2024}, 
distance \cite{chowPhysBenchBenchmarkingEnhancing2025, chenSpatialVLMEndowingVisionLanguage2024,chengSpatialRGPTGroundedSpatial2024}, 
structure \cite{elbananiProbing3DAwareness2024a, zhanGeneralProtocolProbe2024a},
and spatial relations \cite{danierDepthCuesEvaluatingMonocular2025, zhanGeneralProtocolProbe2024a, tongEyesWideShut2024, liuMMBenchYourMultimodal2025, liSEEDBenchBenchmarkingMultimodal2024}.
More explicitly, depth perception was tested by depth ordering of
points \cite{danierDepthCuesEvaluatingMonocular2025, linsley3DPCBenchmarkVisual2025, fuBLINKMultimodalLarge2025},
regions \cite{zhanGeneralProtocolProbe2024a, chengSpatialRGPTGroundedSpatial2024}, or 
objects \cite{chowPhysBenchBenchmarkingEnhancing2025, azadUnderstandingDepthHeight2025, chenSpatialVLMEndowingVisionLanguage2024}.
Notably, the effects of the individual depth cues were examined only in DepthCues \cite{danierDepthCuesEvaluatingMonocular2025}. The authors gathered images from various datasets, labeled them with 6 pictorial depth cues and tested a range of large vision models to show cue utilization. However, since mostly real-world images were used, cues were not isolated. As a result, 60--80\% of images contain occlusion, height-in-plane, relative size and linear perspective cues, whereas other 5 cues appear in fewer than 10\% of the images\footnote{Based on the manual labeling of a random sample of 100 images from \cite{danierDepthCuesEvaluatingMonocular2025}.}.
In contrast, \dsname\ covers a broader set of 9 depth cues via novel synthetic and real images. This offers a more precise control of the pictorial depth cues as well as excludes the possibility of data leakage (\cref{tab:dataset_comp}).

\begin{table*}[bt]
  \caption{Comparison with related datasets. Most existing datasets do not support cue-level analysis, except \cite{danierDepthCuesEvaluatingMonocular2025}. \dsname\ is the only dataset isolating specific cues and cue combinations, allowing direct evaluation of cue utilization.}
  \label{tab:dataset_comp}
  \centering
  \begin{tabular}{@{}lcccrr@{}}
    \toprule
    & \multicolumn{3}{c}{Pictorial depth cues} & & \\
    \cmidrule{2-4}
    Dataset & \# cues & \# present & Controlled & \# images & \# questions \\
    \midrule
    GeoMeter \cite{azadUnderstandingDepthHeight2025} & n/a & multiple & \xmark & 2,000 & 11,200  \\
    3D-PC \cite{linsley3DPCBenchmarkVisual2025} & n/a & multiple & \xmark & 7,574 & n/a \\
    DepthCues \cite{danierDepthCuesEvaluatingMonocular2025} & 6 & multiple & \xmark &  35,753 & n/a \\
    \cmidrule{1-6}
    \dsname\ (ours) & \textbf{9} & \textbf{1 or 2} & \cmark & \textbf{37,110} & \textbf{147,552} \\
    \bottomrule
  \end{tabular}
\end{table*}

\noindent
{\bf Visual question answering (VQA).}
A recent survey \cite{kimVisualQuestionAnswering2025} on VQA stressed the importance of properly using visual information against language bias \cite{niuCounterfactualVQACauseEffect2021} via, for example, weakening language priors or strengthening visual information \cite{goyalMakingVQAMatter2017}.
Studies \cite{dengWordsVisionVisionLanguage2025, chenBenchmarkingRobustnessAdaptation2023} also showed that language had a larger impact on the response of VQA.
While common 3D VQA datasets \cite{azumaScanQA3DQuestion2022, maSQA3DSituatedQuestion2023} contain certain level of question variations, efforts were made to remove unclear and ambiguous questions.
\dsname\ incorporates question variations along many dimensions, including referring clarity (via general \vs specific expressions); together with depth cue variations in \dsname, this allows testing and quantifying vision-language influences on the VQA responses.

\noindent
{\bf Visual grounding.}
The essence of visual grounding is associating language description with corresponding visual stimuli \cite{youFerretReferGround2023, maoGenerationComprehensionUnambiguous2016, pengKosmos2GroundingMultimodal2023, chenScanRefer3DObject2020}. This task is more difficult when multiple objects of a same class are present \cite{liuGRESGeneralizedReferring2023} as wrong associations might impact downstream tasks.
Referring expression comprehension (REC) and phrase grounding are the two main tasks of visual grounding, where the former is concerned with fuller descriptions and the latter focuses on shorter phrases \cite{youFerretReferGround2023}.
A recent work \cite{dahou2025salbench} also explored referring by bounding box coordinates in natural language.
The visual grounding capability is helpful for VQA because it encourages the model to consider visual information \cite{kimVisualQuestionAnswering2025, pengKosmos2GroundingMultimodal2023}. In the proposed \dsname, every image contains multiple same-class objects with \textit{similar appearance}, making visual grounding even more challenging.

\section{Odd-One-Out Depth Dataset}

To systematically evaluate depth perception of VLMs, the proposed \dsname\ dataset incorporates two well-established psychophysical tasks: odd-one-out \cite{sinapovOddOneOut2010} and depth ordering \cite{brennerDepthPerception2018}. Each image in the \dsname\ contains multiple similar objects, with only one object (the target) located at a different depth plane relative to other objects (the distractors). To formulate it as a Visual Question Answering (VQA) task, we design a set of prompts for each image with varying referring clarity. Referring expression comprehension \cite{liuGRESGeneralizedReferring2023} allows testing the language abilities of the models. A detailed description of our methodology follows with additional information available in the Supplementary Material.

\noindent
{\bf Pictorial depth cues}. To analyze depth perception, we use the following 9 common pictorial cues identified in the psychology literature \cite{reicheltDepthCuesHuman2010, kavsekInfantsSensitivityPictorial2012, wattFocusCuesAffect2005}: Height-in-Plane (HP), Occlusion (OC), Relative Size (RS), Familiar Size (FS), Light-and-Shadow (LS), Texture Gradient (TG), Aerial Perspective/Saturation (SA), Focusness (FO), and Linear Perspective (LP). 

\begin{figure}[tb]
  \centering
  \begin{minipage}[b]{0.315\textwidth}
      \begin{subfigure}{\textwidth}
      	\includegraphics[width=\textwidth]{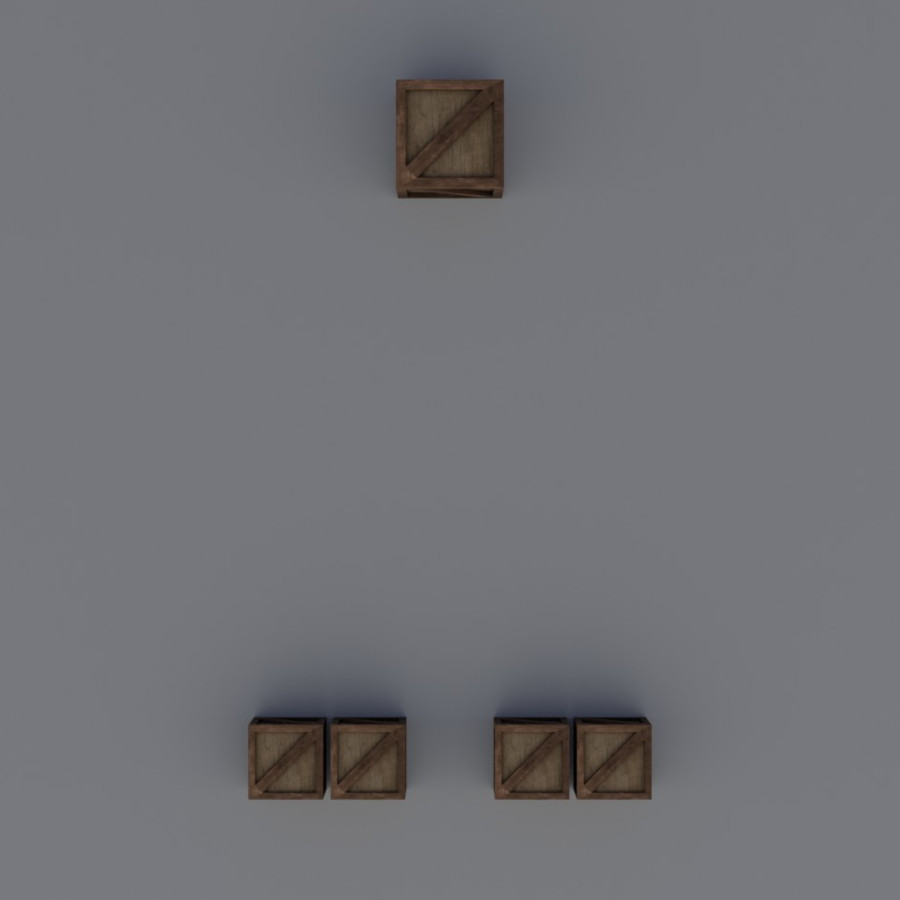}
            \caption{Bird's-eye view}
            \label{fig:base-bev-a}
      \end{subfigure}
      \begin{subfigure}{\textwidth}
      	\includegraphics[width=\textwidth]{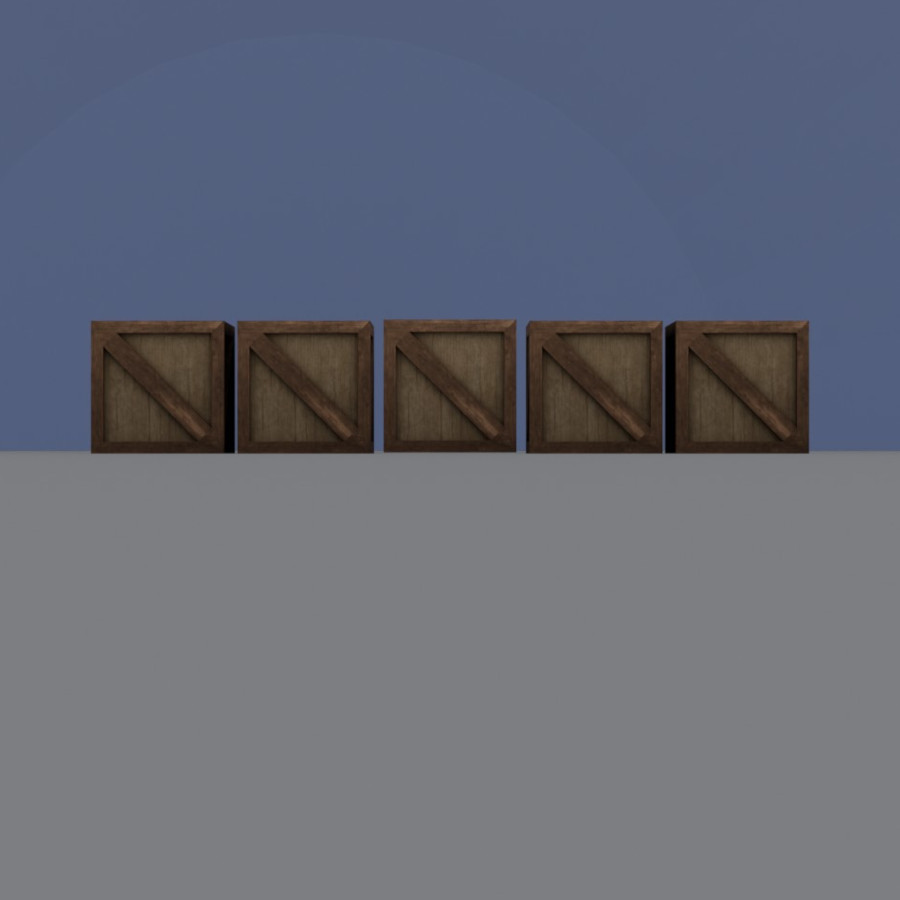}
            \caption{Ambiguous (base) view}
            \label{fig:base-bev-b}
      \end{subfigure}
  \end{minipage}
  \hfill
  \begin{minipage}[b]{0.665\textwidth}
      \begin{subfigure}{\textwidth}
      	\includegraphics[width=\textwidth]{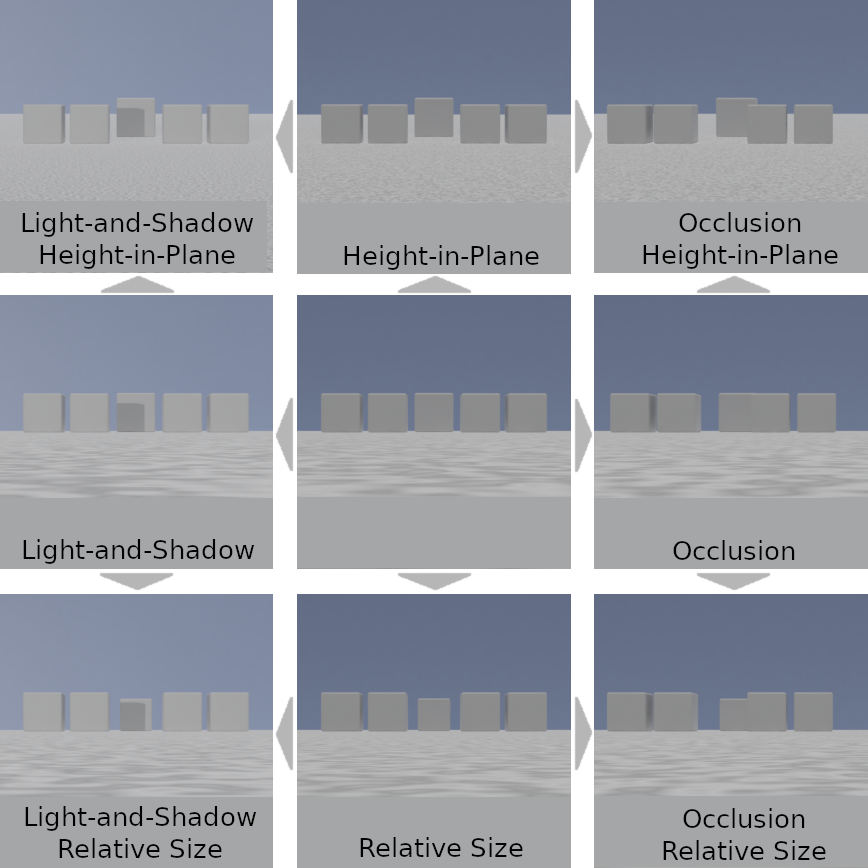}
            \caption{Views with depth cues derived from the base view}
            \label{fig:base-bev-c}
      \end{subfigure}
  \end{minipage}
  \caption{Each 3D scene in \dsname\ contains 1 target and 4 distractors, where the target is larger (or smaller) and located on a different depth plane (\protect\subref{fig:base-bev-a}). When a camera is placed at a certain position, (\protect\subref{fig:base-bev-b}) the target appears at the same depth as the distractors. (\protect\subref{fig:base-bev-c}) Gray arrows between the images indicate how disambiguated 2D views are generated from the base view (in the center) by adding one or two pictorial depth cues.}
  \label{fig:base-bev}
\end{figure}

\noindent
{\bf Scene construction \& cue control}.
Each \dsname\ scene contains 1 target and 4 distractors placed on a level surface. The target differs from the distractors only in size and is placed at a different depth plane from the distractors (\cref{fig:base-bev-a}).
Positioning the camera in a certain way (\cref{fig:base-bev-b}) creates views with scale ambiguity \cite[p.~53]{szeliskiComputerVisionAlgorithms2022}.
It is emphasized that scene and view are 3D and 2D concepts, respectively. In other words, from a single 3D scene we generate multiple 2D views with different depth cues. See Fig 1. in Supplementary Material for examples of each.

\cref{fig:base-bev-c} illustrates the cue control method and shows some resulting views.
From an ambiguous base view (\cref{fig:base-bev-c}, center), individual depth cues are added by manipulating the camera, objects, or environment.
Translating the camera horizontally and upward introduces the OC and HP cues, respectively; moving the target along the optical axis results in the RS cue;
adding directional light so that near objects cast shadows on far ones gives the LS cue;
adding textures to objects creates the TG cue;
removing textures from the ground controls the LP cue \cite{reicheltDepthCuesHuman2010};
using a larger camera aperture strengthens the FO cue;
common objects with similar shapes but different sizes are used for the FS cue;
finally, the natural source of the SA cue is haze, which we simulate by a haze equation \cite{heSingleImageHaze2009}.
While \cref{fig:base-bev-c} only shows 4 single-cue and 4 two-cue views, \dsname\ dataset covers 9 individual cues as well as all possible second-order interactions among them\footnote{Except Linear Perspective (LP) which has to be tested with Height-in-Plane (HP) because both cues are the result of moving camera above ground.}.

\begin{table}[tb]
  \caption{Image and question counts in \dsname\ dataset, grouped by number of pictorial cues present in the images.}
  \label{tab:o3d_img_ques}
  \centering
  \begin{tabular}{lrr@{\hspace{5pt}}r@{\hspace{5pt}}l}
    \toprule
    Cues & Real imgs & Sim imgs & Img-ques pairs & Reason to include \\
    \midrule
    0-cue  &  16  & 1,443 & 5,796 & as negative baseline \\
    1-cue & 99 & 13,746 & 55,170 & controlled single-cue images\\
    2-cue & 79 & 21,556 & 86,244 & controlled two-cue images\\
    mixed & 171 & - & 342 & as natural baseline\\
    \midrule
    total & 365 & 36,745 & 147,552 & \\
    \bottomrule
  \end{tabular}
\end{table}

\noindent
{\bf Synthetic scenes.} We use the Kubric \cite{greffKubricScalableDataset2022} simulation environment to render a set of images with the following characteristics:
\begin{itemize}
    \item the target is randomly scaled to be 10\% to 100\% larger (smaller), and placed on a farther (nearer) depth plane relative to distractors;
    \item 37 object classes selected from Kubric assets, with different shape complexities, ranging from simple boxes to complex toys;
    \item 13 environments with diverse indoor and outdoor backgrounds and realistic ambient lighting;
    \item 9 individual pictorial cues and 28 pairs of cues. For each cue except FS and LP, we additionally generate images with various cue strengths. For example, cue strengths of HP and OC can be measured by height difference and area of occlusion, respectively.
\end{itemize}
\noindent
Overall, we render 13,746 images with single cue and 21,556 with two-cue interactions (see \cref{tab:o3d_img_ques}).
Also obtained from the Kubric renderer are depth maps, segmentation maps, as well as ground truth target and distractors. More information on the Kubric setup is in Section 2 of Supplementary Material.

\noindent
{\bf Real-world scenes.} We set up real-world \dsname\ scenes in an indoor environment.
Following the same procedure as in simulation, multiple views are derived and captured for 3 object classes and 8 cues (excluding Saturation).
The 3 object classes are cube, clip, and cup, with the targets being 100\%, 25\%, and 27\% larger, respectively.
We use a dark-colored desk as the level surface and a green wall as the background.
The above forms our 1-cue and 2-cue real images, as summarized in \cref{tab:o3d_img_ques}.

\noindent
{\bf Sensor settings \& post-processing.} In both simulation and real world, the camera focal length is set close to 50 mm. When the FO cue is unwanted, we turn off depth of field rendering in Blender, or use a small aperture (\eg f/18). We try to reduce undesired variations between shots: for each scene, auto-exposure is run once and turned off; focus is manually set on the center object. The rendered image size is 1024 $\times$ 1024, while real images were resized to 1024 $\times$ 683.

\noindent
{\bf Image baselines.}
\dsname\ contains two special subsets of images: 0-cue and mixed-cue (\cref{tab:o3d_img_ques}).
The 0-cue set consists of all the ambiguous (base) view images (\eg \cref{fig:base-bev-b}) with no pictorial cues.
For the mixed-cue set, we select 171 real-world  images from O$^3$ dataset \cite{kotserubaSaliencyModelsDetect2021} where the odd target was behind or in front of all the distractors.
In these images, multiple pictorial cues exist and are not controlled. We label them with two most prominent cues for reference. O$^3$ images were resized to a max of 1024 pixels in the larger dimension.

\begin{table}[tb]
  \caption{Referring expression variations. As referring to the target object in an \dsname\ image is challenging, we explore referring expressions with different clarity.}
  \label{tab:ref_var}
  \centering
  \resizebox{0.8\textwidth}{!}{
  \begin{tabular}{c@{\hskip 1em}c@{\hskip 1em}l}
    \toprule
    Feature & Clarity & Example expression \\
    \midrule
    saliency & low &Is the \textit{salient object} ... \\
    depth ($z$) & med & Is the \textit{salient object due to its unique distance} in the image ... \\
    spatial ($x,y$) & high & Is the \textit{object in the middle} ... \\
    marker & highest & Is the \textit{object marked with a red circle} ... \\
    \bottomrule
  \end{tabular}
  }
\end{table}

\begin{figure}[b]
  \centering
  \begin{subfigure}{0.35\linewidth}
  	\includegraphics[width=\linewidth]{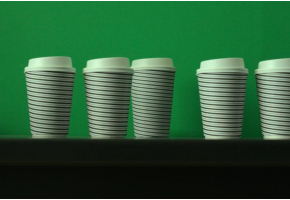}
        \caption{Resized}
        \label{fig:amc-a}
  \end{subfigure}
  \begin{subfigure}{0.35\linewidth}
  	\includegraphics[width=\linewidth]{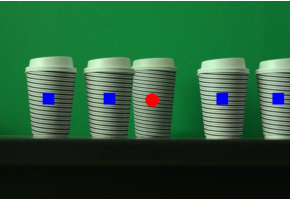}
        \caption{Marked}
        \label{fig:amc-m}
  \end{subfigure}
  \caption{Image post-processing.
  (\protect\subref{fig:amc-a}) Resize a real image to 1024 $\times$ 683.
  (\protect\subref{fig:amc-m}) Optionally add markers for easier referring (\cref{tab:ref_var} bottom row), to gauge the effects of referring expression comprehension (REC) on depth ordering responses.
  }
  \label{fig:aug_mark_crop}
\end{figure}

\noindent
{\bf Prompt templates.} We generate a template-based question space of 1,026 unique prompts, with variations in 4 dimensions: query template, target referring, distractor referring, and response instruction. 

Specifically, we obtain 9 templates by collecting depth questions from common VQA datasets \cite{linsley3DPCBenchmarkVisual2025, chowPhysBenchBenchmarkingEnhancing2025, chengSpatialRGPTGroundedSpatial2024, chenSpatialVLMEndowingVisionLanguage2024, maSQA3DSituatedQuestion2023}, and rephrasing them with an LLM \cite{Shah_2019_CVPR}. The query templates address various perspectives of depth ordering questions, such as different vocabularies, regular \vs yes-no queries, and levels of formalism (see Section 4 in Supplementary Material).
Each template also contains two placeholders for target and distractor referring expressions.

As shown in \cref{tab:ref_var}, the target referring expressions vary by their clarity.
We explore 4 levels of clarity with distinct referring features.
The referring expressions of low and medium clarity require understanding the target as an odd object, described by \eg ``salient'' or ``standing out''.
Because REC in a context with multiple similar objects is challenging \cite{chengSpatialRGPTGroundedSpatial2024, liuGRESGeneralizedReferring2023}, we additionally test a more direct referring mechanism with visual markers placed on objects (\cref{fig:amc-m}). Our referring expressions across clarity levels cover all 3 common REC variations of subject, location, and relation \cite{QiaoRECSurvey2021}.

To simplify VLM response parsing, we use a multiple-choice question format, as in \cite{azadUnderstandingDepthHeight2025, linsley3DPCBenchmarkVisual2025, fuBLINKMultimodalLarge2025}.
Since only one object (\ie the target) is at a different depth, the depth order can be described with binary answers (\eg closer or farther). Therefore, the response instruction is simply a 2-item option list (\eg \textit{A. farther. B. closer.}) followed by an instruction (\ie \textit{Answer A or B.}).
The option list is randomized, resulting in normal and reversed orders.
We use the forced binary choice because preliminary results showed that if a `not sure' option was included, VLMs almost always chose it, thus making the analysis less meaningful.

\textbf{Prompts.} For each image in \dsname, we sample \textit{depth ordering} questions with different target referring clarity from the prompt templates. This results in 147K image-question pairs (\cref{tab:o3d_img_ques}).

\section{Experiment Setup}
We run 12 VLMs on \dsname\ and evaluate their VQA performance on various question formats and pictorial cues.

\noindent
{\bf Baseline.} DepthAnythingV2 \cite{yangDepthAnythingV22024} is selected as the baseline for its robustness to diverse scenes and ability to process high-resolution images.
Since DepthAnythingV2 produces depth maps only, we further process the output to make comparisons with VLMs.
Specifically, we compute the median depth of target and distractors using their ground truth masks, then determine the depth ordering.

\noindent
{\bf Evaluated VLMs.}
We evaluate the following VLMs:
Kosmos2 \cite{pengKosmos2GroundingMultimodal2023},
PaliGemma2 \cite{steinerPaliGemma2Family2024},
Qwen2-VL \cite{wangQwen2VLEnhancingVisionLanguage2024},
InternVL2.5 \cite{chenExpandingPerformanceBoundaries2025},
DeepSeek-VL \cite{luDeepSeekVLRealWorldVisionLanguage2024},
LLaVA1.5 \cite{liuImprovedBaselinesVisual2024},
VILA1.5 \cite{linVILAPretrainingVisual2024},
BLIP2 \cite{liBLIP2BootstrappingLanguageImage2023},
Phi3 \cite{abdinPhi3TechnicalReport2024},
Cambrian \cite{tongCambrian1FullyOpen2024},
GPT4.1-mini, and Gemini 2.5 Flash-Lite.
All evaluated VLMs are open-source, except GPT and Gemini.
Out of these VLMs, only the Cambrian model particularly focuses on utilizing visual information.
In addition to language referring, Kosmos2 supports region referring by bounding boxes; results from both referring types will be reported. 

Depth-focused SpatialRGPT \cite{chengSpatialRGPTGroundedSpatial2024} is not selected because it takes mask referring as part of the input and runs DepthAnything in the background, which is equivalent to our baseline described above.

\noindent
{\bf Metrics.} We measure accuracy for all responses. In addition, we introduce the standard deviation of within-group means (SDGM) metric (see \cref{sec:lang_dim_results}) to measure VLMs' sensitivities to cue and language variations in the depth ordering task.

\noindent
{\bf In-context learning (ICL) and chain-of-thought (CoT) prompting.}
For 5 of 12 VLMs, we provide additional few-shot ICL \& CoT prompting using the 1-cue, 2-cue, and mixed-cue images (\cref{tab:o3d_img_ques}). As image similarity and order matters \cite{Baldassini_2024_CVPRW}, we retrieve two (target-far and target-near) demonstrations with the same cues as in the main image, in randomized order. Within each demonstration, the image is positioned before the depth question prompt followed by the expected answer \cite{Qin_NEURIPS2024_deeb4d6b}. The CoT prompting in the demonstrations only addresses the depth understanding, not the referring comprehension. As an example, a demonstration can be formatted as follows: 
\begin{quote}
<image> Is the unique object positioned farther from or closer to the observer than the remaining objects? A. Farther. B. Closer. Answer A or B. (Let's think step by step. The object of interest appears lower than the others.  Based on the height-in-plane pictorial cues, it is likely that the object is closer than the other objects.) B.
\end{quote}

\noindent
{\bf Response parsing.}
We parse only the first sentence\footnote{CoT prompting occasionally caused extra outputs, which we removed before parsing.} in a VLM response.
Most responses are simply \textit{A} or \textit{B}, and we report accuracy against the ground truth.
Responses that do not follow the formatting instruction are scored as positive only if their first sentence contains the correct answer option.

\begin{figure}[tb]
  \centering
  \includegraphics[height=6.5cm]{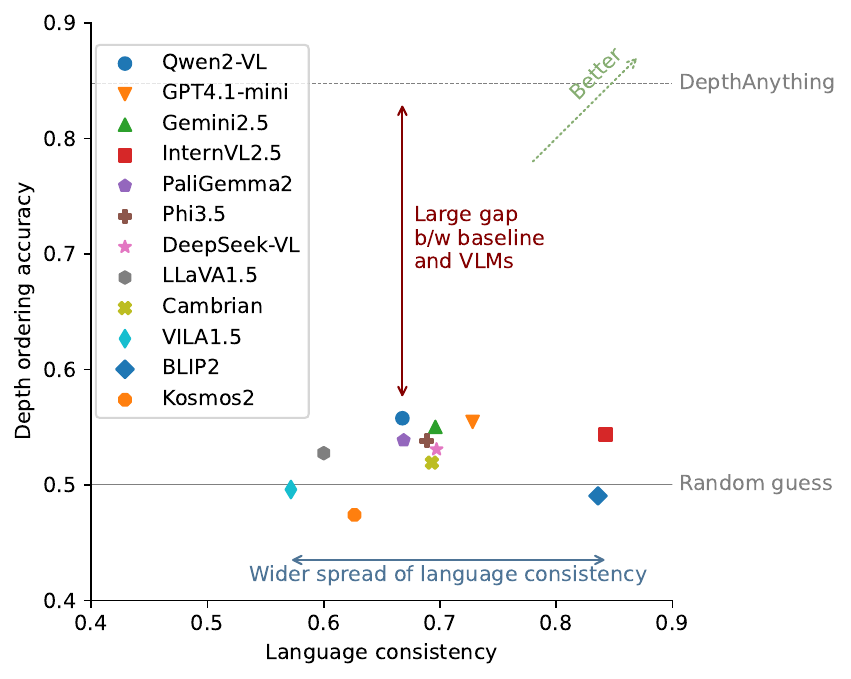}

   \caption{Performance summary of VLMs on depth ordering VQA.
   Both depth understanding (y-axis, see \cref{sec:depth_dim_results}) and language consistency (x-axis, see \cref{sec:lang_dim_results}) can be probed using our \dsname\ dataset.
   Depth ordering accuracies of VLMs are close to random guess and inferior to DepthAnythingV2 \cite{yangDepthAnythingV22024} baseline.
   VLMs' language consistency has a wide spread.}
   \label{fig:d_l_scores}
\end{figure}

\section{Experiment Results}
As summarized in \cref{fig:d_l_scores}, we report the results of experiments that probed VLMs' understanding of pictorial depth cues (\cref{sec:depth_dim_results}) and question comprehension (\cref{sec:lang_dim_results}).
In addition, we discuss cue-level findings and a bias along the near-far spectrum for the vision dimension, and address the consistency of VQA responses for the language dimension. Additional experiment results are presented in Section 6 of Supplementary Material.

\subsection{Vision Dimension}
\label{sec:depth_dim_results}
\begin{figure}[t]
  \centering
  \includegraphics[height=8cm]{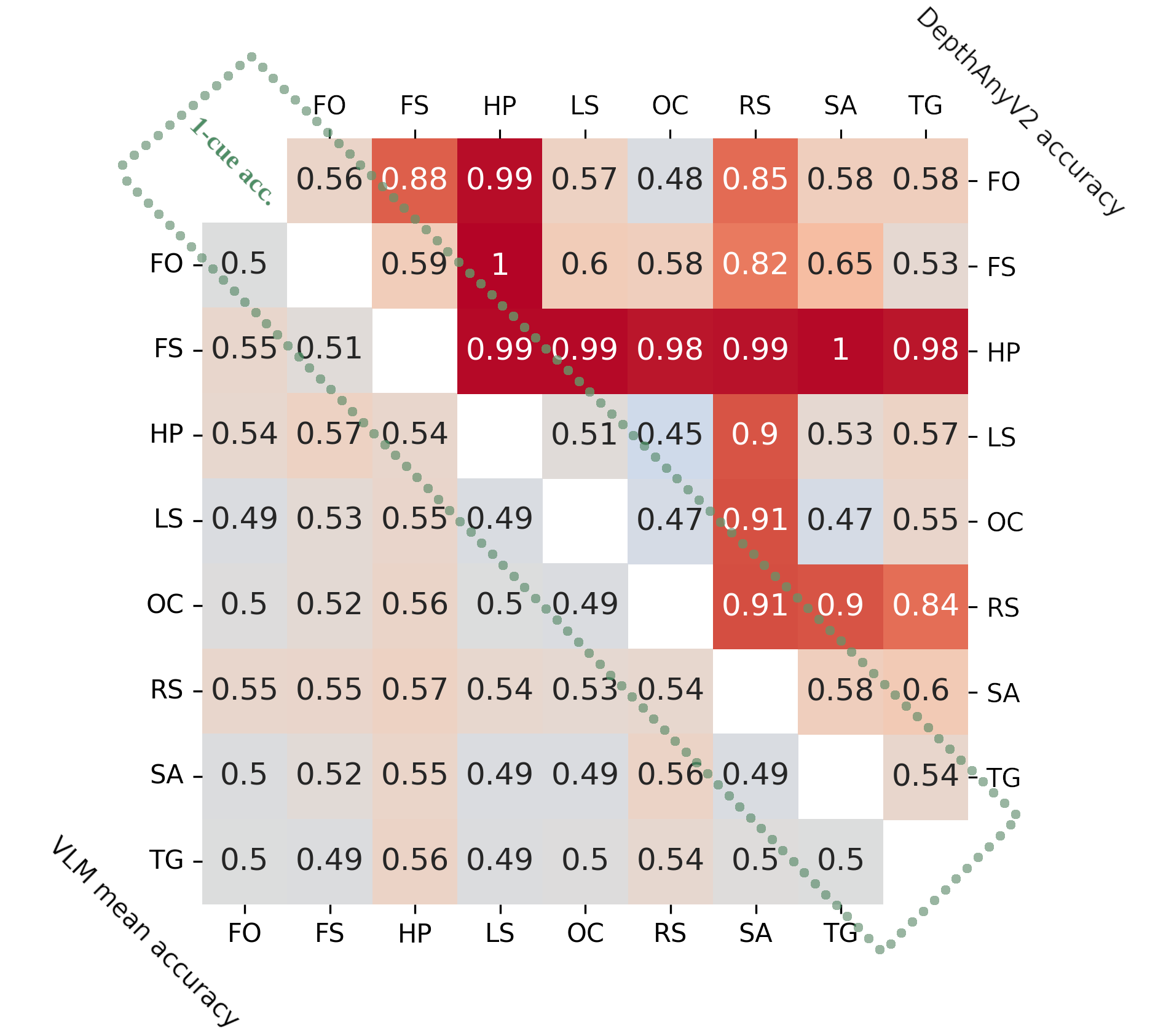}
   \caption{A combined heatmap of 1-cue and 2-cue mean accuracies of all tested VLMs (bottom-left), \vs baseline (top-right). Red- and blue-tinted cells indicate performance above and below chance level (0.5).
   The two main diagonal cells (within green dotted rectangle) show accuracies for 1-cue depth ordering, whereas the other cells report 2-cue interactions.
   The depth ordering performance is better whenever HP or RS cue are present.
   (FO: Focusness, FS: Familiar Size, HP: Height-in-Plane, LS: Light-and-Shadow, OC: Occlusion, RS: Relative Size, SA: Saturation, TG: Texture Gradient)
   }
   \label{fig:cue_heatmap}
\end{figure}

In order to reduce the interference of referring expression comprehension, the results of vision dimension are reported only for the images with markers, \ie the highest referring clarity in \cref{tab:ref_var}, unless otherwise noted.
When comparing with 2-cue results (\cref{fig:cue_heatmap,tab:dep_dim_baselines}), we ensure the same cue strength across the 1-cue and 2-cue images.

\noindent
{\bf VLMs perform at chance level for single pictorial cues and 2-cue combinations.}
We find a performance gap (\cref{fig:d_l_scores,fig:cue_heatmap}) between all tested VLMs and the DepthAnythingV2 baseline in terms of depth ordering accuracy.

\cref{fig:cue_heatmap} is a cue-level heatmap of mean VLM accuracy (lower triangle) and DepthAnythingV2 baseline accuracy (upper triangle). Red-colored cells indicate better performance. The main diagonal cells (in the dotted rectangle) of the two sub-heatmaps report 1-cue accuracies, and the other cells are for 2-cue.
As shown by the lower main diagonal, the VLMs perform the best on Height-in-Plane (0.54), Relative Size (0.54) and Familiar Size (0.51), although only marginally above the chance level (0.5).
The largest performance gap between the VLMs and the baseline are also from the best cues of HP ($\delta=0.45$) and RS ($\delta=0.37$).

As one would expect, a combination of depth cues improves human depth perception \cite{westermanIndividualDifferencesUse1998}.
For VLMs, we observe such trend but only to a limited extent, shown by the 2-cue numbers (under the main diagonal in \cref{fig:cue_heatmap}) being slightly greater than the 1-cue accuracies.
The depth ordering are more accurate whenever HP or RS cue is present. Most other 2-cue combinations have insignificant influence on VLMs.
As summarized in \cref{tab:dep_dim_baselines}, both VLMs and DepthAnythingV2 perform better when more cues are present.

The model-level accuracy gaps are similar to the aggregated means in \cref{fig:cue_heatmap} (see Supplementary Material). None of the evaluated VLMs perform significantly better than chance; the commercial (GPT4.1-mini \& Gemini2.5) and vision-centric (Cambrian) models are no exceptions.

\begin{table}[tb]
  \caption{Depth ordering accuracy comparing with model \& image \textit{baselines}.
  We run DepthAnythingV2 on all images as a reference. 
  DepthAnythingV2 has improved performance for real images and when more cues present, whereas VLMs have a similar pattern but much weaker.
  }
  \label{tab:dep_dim_baselines}
  \centering
  \begin{tabular}{@{}lrr@{}}
    \toprule
    Image set & VLM mean & \textit{DepthAnyV2} \\
    \midrule
    \textit{real (mixed cues)} & 0.6051 & 0.9375 \\
    real (1-cue \& 2-cue) & 0.5392 & 0.9660 \\
    sim (1-cue \& 2-cue) & 0.5221 & 0.7090 \\
    \midrule
    \textit{0-cue} & 0.4865 & 0.5177 \\
    1-cue & 0.5165 & 0.6535 \\
    2-cue & 0.5282 & 0.7446 \\
    \bottomrule
  \end{tabular}
\end{table}

\noindent
{\bf Linear Perspective (LP) cue interacts positively with other cues.}
Based on results in DepthCues \cite{danierDepthCuesEvaluatingMonocular2025}, one of the best cues for depth perception was LP. We test it in a negative way, where the LP cue is eliminated from a view by removing ground texture \cite{reicheltDepthCuesHuman2010}. In addition to confirming the claims in DepthCues \cite{danierDepthCuesEvaluatingMonocular2025}, our results show LP has mostly positive interactions with other cues. The average accuracy gain due to LP is about 0.03, matching the best cues Height-in-Plane (HP) \& Relative Size (RS).

Following \cite{danierDepthCuesEvaluatingMonocular2025}, we apply Spearman correlation to the 8 single-cue depth ordering accuracies of the evaluated VLMs, but obtain inconsistent results (see Supplementary Material).
While RS/HP had high correlation among cue tasks in both \cite{danierDepthCuesEvaluatingMonocular2025} (0.82) and our analysis (0.60), another highly correlated pair of RS/OC (0.75) in \cite{danierDepthCuesEvaluatingMonocular2025} is nearly uncorrelated (0.07) in our results. Our overall correlations among different cues for depth ordering are lower, a sign of more controlled pictorial cue analysis.

\noindent
{\bf In-context learning and chain-of-thought prompting helps commercial VLMs only, with limited improvements.}
Among the 5 VLMs tested with ICL and CoT prompting, only the commercial GPT4.1-mini ($\delta=0.068$) and Gemini2.5 ($\delta=0.042$) benefit from it. The additional few-shot demonstrations hinder the performance of DeepSeek ($\delta=-0.022$), Qwen2 ($\delta=-0.012$), and Phi3.5 ($\delta=-0.006$). For all 5 VLMs, however, the performance differences are not significant across cues. The ICL and CoT prompting is most effective for the RS cue ($\delta=0.069$) and least for FS ($\delta=-0.036$).

\begin{figure}[tb]
 \centering
 \includegraphics[height=7cm]{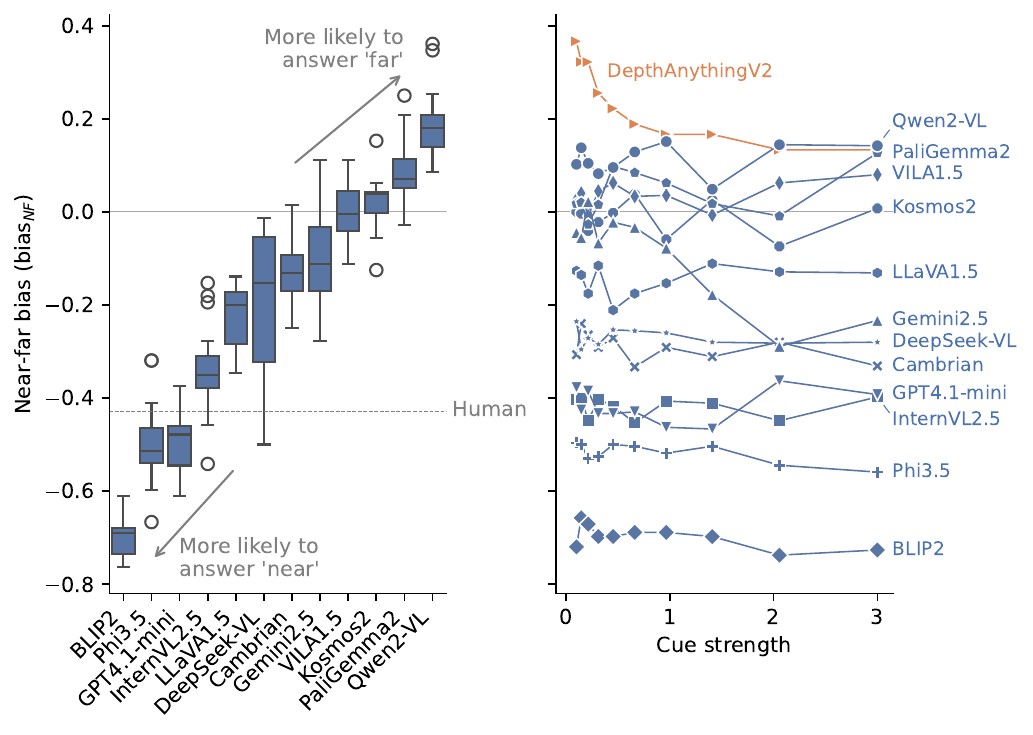}

  \caption{Near-far bias. Both plots share the $y$-axis. Left: Boxes show variations across cues. VLMs exhibit different extents of biases toward answering near \vs far, compared with human reference \cite{chenSingleImageDepthPerception}. 
  Right: None of the VLMs has consistent bias reduction as cue strength increases, compared with DepthAnythingV2.
  }
  \label{fig:nf_bias}
\end{figure}

\noindent
{\bf VLMs have a wide range of biases towards answering near \vs far.}
Most image sets in \dsname\ (\cref{tab:o3d_img_ques}) are class-balanced, meaning that the number of images with the target located near is equal to that of far.
This near-far balanced property allows us to explore VLMs' biases towards near \vs far when answering depth ordering questions (\cref{fig:nf_bias}).

In this paper, the near-far bias of a VLM is defined as 
\begin{equation}
  \mathrm{bias}_{NF}(M) = \mathrm{Acc}_F(M) - \mathrm{Acc}_N(M),
  \label{eq:nf_bias}
\end{equation}
where $\mathrm{Acc}_N$ and $\mathrm{Acc}_F$ denotes the accuracy of a model $M$ on the target-near and target-far image subsets, respectively.
We argue without proof that, given near-far balanced dataset and randomized questions (see \cref{sec:lang_dim_results}), $\mathrm{bias}_{NF}$ should reflect VLMs' preferences on answering near \vs far.
Namely, $\mathrm{bias}_{NF}$ should be zero when a VLM is unbiased.
If a VLM always answers ``far" or ``near'' ignoring vision inputs, $\mathrm{bias}_{NF}$ will be $1.0$ or $-1.0$, respectively.

\cref{fig:nf_bias} reports $\mathrm{bias}_{NF}$ (left panel) and its change (right panel) with increased cue strengths for each benchmarked VLM.
The VLMs have a wide spread of $\mathrm{bias}_{NF}$ ranging from $-0.7$ to $0.2$.
Majority of the VLMs prefer answering near (below the 0.0 line). BLIP2, Phi3, and GPT4.1-mini are the three most near-biased VLMs, whereas Qwen2-VL and PaliGemma2 are far-biased. 
The human baseline of  $-0.428$ is computed based on \cite{chenSingleImageDepthPerception}, whose authors found a positive correlation between center pixels and nearness for human judgments on web collected images. The baseline is comparable because we always have the target close to the image center and the distractors more eccentric.

We also analyze whether increasing cue strengths will reduce the near-far bias, and the answer is negative, as shown in the right panel of \cref{fig:nf_bias}.
Once again, the difference between VLMs and the baseline is apparent. VLM biases (shown as blue lines) do not converge towards 0 as cues become stronger. DepthAnythingV2 (orange line), in contrast, has a clear trend of bias reduction.

\subsection{Language Dimension}
\label{sec:lang_dim_results}

Intuitively, if a VLM understands depth ordering in an image, the responses should remain consistent for any equivalent questions.
In other words, better models should have higher language consistency and less variation.
To measure this, we introduce a variation-based metric. The standard deviation of within-group means (SDGM) is given by
\begin{equation}
  \sigma_{\Omega}(\mu) =  \sqrt{\frac{1}{||\Omega||} \sum_{g \in \Omega}(\mu_g - \bar{\mu})^2},
  \label{eq:SDGM}
\end{equation}
where $\Omega$ defines a set of groups, and $\mu_g$ denotes a mean performance metric within each group $g$.
If, for example, $\Omega$ is the target referring clarity in \cref{tab:ref_var}, there will be 4 groups, and 4 within-group means $\{\mu_{g_i}\}^4_{i=1}$. Then we can obtain SDGM by computing the standard deviation of the means.
For simpler analysis, we use the target-near accuracy in \cref{eq:nf_bias} as the metric, \ie $\mu := \overline{Acc_N}$.

A good model should yield a low SDGM, the variation metric, \emph{only when the groups in $\Omega$ are equivalent} in terms of model performance.
Our question variations do not alter the essence of the depth ordering questions, so they satisfy this condition. Therefore, out of the 1,026 unique prompts, we define question groups $\Omega_{L}$ by tagging the prompts from 5 perspectives, ending up with 42 groups. For example, one group may have the tags: \texttt{high} clarity, \texttt{close-far} vocabulary, \texttt{formal}, \texttt{yes-no} query, and \texttt{normal} order. With a minor modification, we can use SDGM to analyze language consistency for a specific dimension of question variation, \eg \texttt{normal} \vs \texttt{reversed} order (see Supplementary Material).
Finally, the language consistency metric in \cref{fig:d_l_scores} is given by $C = 1 - \sigma_{\Omega_{L}}(\overline{Acc_N})$. 

We also exploit SDGM more generally, using it to represent the influence of the grouping criteria on the model performance. In other words, the SDGM of a model will be high if the model is (over-)sensitive to the differences among groups.
To compare language influence with vision, we define cue groups $\Omega_V$ consisting of the 8 single-cue and 28 two-cue cases. The total 36 cue groups is comparable to the 42 question groups when computing SDGM.

\begin{figure}[tb]
  \centering
  \includegraphics[height=5cm]{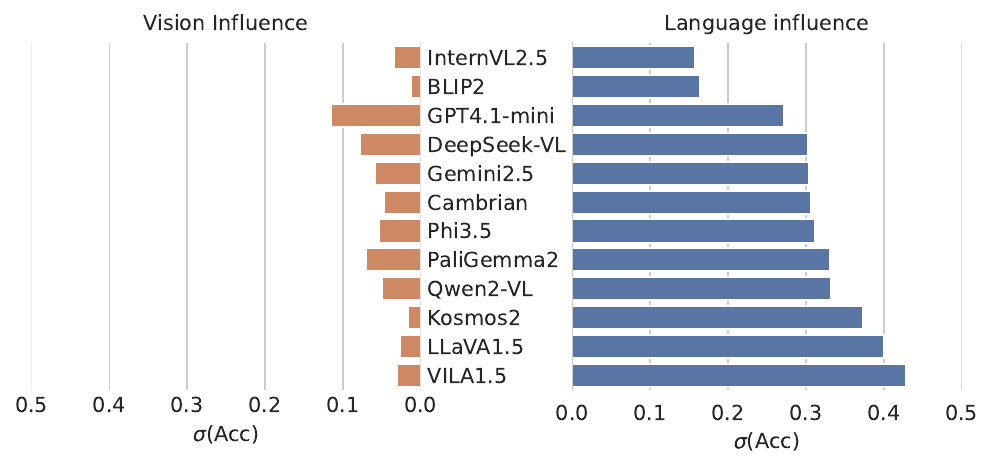}

   \caption{Vision \vs language influence on depth ordering VQA.
   Bars show the standard deviations of mean accuracy (SDGM, see text), as a measure of influence.
   Language influence is uniformly larger than vision.
   }
   \label{fig:vl_influ}
\end{figure}

\noindent
{\bf Language has a much larger influence on VQA responses than vision.}
\cref{fig:vl_influ} reports the quantified vision ($\sigma_{\Omega_V}$) and language ($\sigma_{\Omega_L}$) influences for each VLM. The language influence is uniformly larger than that of vision. InternVL2.5 has the least blind faith in text \cite{dengWordsVisionVisionLanguage2025} and the smallest difference between vision and language. The high language sensitivity of the VLMs is consistent with results in \cite{chenBenchmarkingRobustnessAdaptation2023}.

\noindent
{\bf Inconsistencies are mainly caused by varying depth order vocabulary and \texttt{yes-no}/\texttt{multi-choice} query.}
Depth order vocabulary has three tags: \texttt{close-far}, \texttt{front-rear}, and \texttt{before-after}, which describes the depth relation between the target and distractors using different pairs of words. This source of variations causes the largest inconsistency ($\sigma=0.2166$).

Whether a question is yes-no (\textit{Is it behind?}) or multi-choice (\textit{Is it closer or farther?}) also gives rise to a large undesired performance deviation ($\sigma=0.1931$).
InternVL2.5, as discussed above, is significantly more consistent against the yes-no query variation. Most VLMs are biased towards answering positive to the yes-no queries compared with the more neutral multi-choice queries. DeepSeek-VL and PaliGemma2 are two exceptions in that they answer ``no'' more often, which could be a result of over-correction (see Supplementary Material).

\noindent
{\bf The \texttt{normal}/\texttt{reversed} order variation affects just a few VLMs.}
We tag a prompt as \texttt{reversed} order when the response options are reversed from the expression in the question (\eg \textit{Is it closer or farther? A. farther. B. closer.}).
The most negatively impacted VLMs are VILA1.5 ($\sigma=0.5608$) and Qwen2-VL ($\sigma=0.3719$), followed by LLaVA1.5. This is in line with similar findings for large language models \cite{pezeshkpourLargeLanguageModels2023}.
The remaining VLMs have a similar level of robustness against different option orders (see Supplementary Material).

\noindent
{\bf Referring clarity slightly helps depth ordering.}
As referring clarity (\cref{tab:ref_var}) increases, the overall depth ordering accuracy slightly improves for both synthetic and real-world images. Adding markers results in the largest but still marginal improvement.
Similarly, referring with bounding boxes by Kosmos2 brings a negligible improvement (see Supplementary Material). This suggests that referring expression comprehension is not a major hurdle, rather depth is.

\section{Conclusion}
In this work, we introduced \dsname, a dataset for testing VLMs and vision models on depth ordering.
To the best of our knowledge, our work was the first systematic investigation of VLM performance across a set of pictorial cues, thanks to the novel scene construction and cue control method. Language complexity was also addressed, both independently and in combination with vision.
Our experiments showed that all evaluated open-source and commercial VLMs performed around the chance level, regardless of utilized depth cues, their cue strengths, level of visual realism, and referring clarity.
We quantified vision \& language influences on the VQA responses, and found that the VLMs were significantly more sensitive to language input than vision.
These results have implications for applications involving VLMs, such as robotic manipulation and human-robot interaction.
Major future work includes extending current focus on pictorial cues to motion-based monocular cues, and further to binocular ones.
We hope that the data and evaluation approach presented in this paper will help bridge the gap between language and vision components of VLMs, as well as bring machine vision closer to its biological counterpart.

\section*{Acknowledgements}
This work was supported by grants to JKT from the Natural Sciences and Engineering Research Council of Canada (NSERC) under award number RGPIN-2022-04606 and the Air Force Office for Scientific Research (USA) under award number FA9550-22-1-0538.

%
%
\bibliographystyle{splncs04}
\bibliography{main,oood}

\bookmarksetup{startatroot}
\title{Disentangling Pictorial Cue Understanding from Language Bias in VLMs via Depth Ordering Task (Supplementary Material)}

\author{}
\institute{}

\maketitle

\section{Controlled Pictorial Cues}

\begin{figure}
    \centering
    \includegraphics[width=\linewidth]{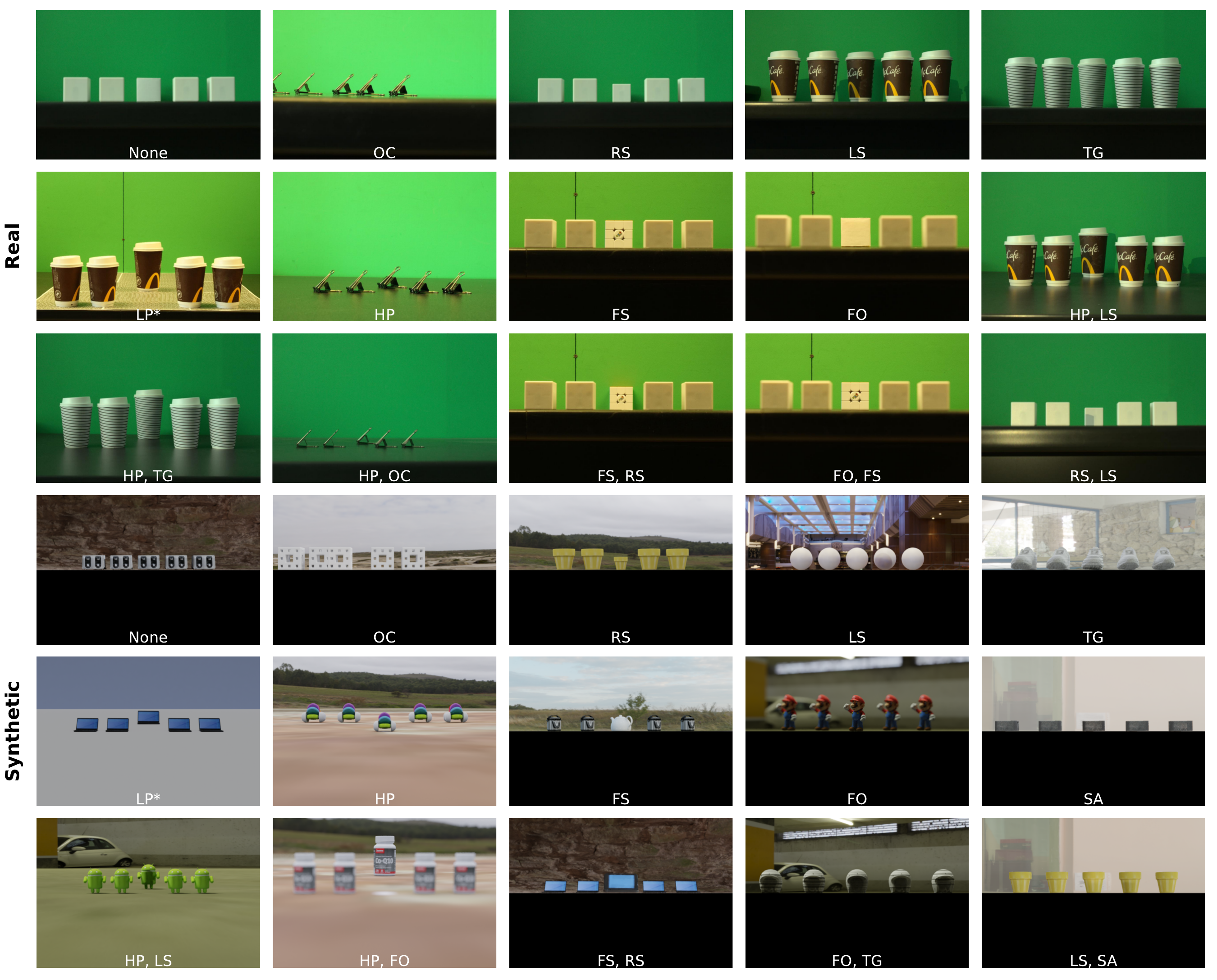}
    \caption{Real and synthetic images with controlled cues in \dsname. Specially, the LP cue (leftmost image in 2nd and 5th row) has to be tested with the HP cue in a relative way: we control the LP by manipulating ground textures \cite{wickensThreeDimensionalDisplaysPerception1989}.}
    \label{fig:common-pic-cues}
\end{figure}

\cref{fig:common-pic-cues} illustrates sampled images for the base view and 9 controlled pictorial cues: Occlusion (OC), Relative Size (RS), Light-and-Shadow (LS), Texture Gradient (TG), Linear Perspective (LP), Height-in-Plane (HP), Familiar Size (FS), Saturation (SA), and Focusness (FO). Images with no cues (leftmost in the 1st and 4th row) and two cue combinations are also included.

Without some special conditions, the Focusness (FO) cue alone cannot provide relative depth ordering \cite{Marshall1996, wattFocusCuesAffect2005}. This can be confirmed by the fact that FO accuracies in \cref{tab:vlms_v} are similar to 0-cue. We, therefore, include FO mostly for 2-cue interactions.

Testing Linear Perspective (LP) requires a special approach. According to \cite{wickensThreeDimensionalDisplaysPerception1989}, the presence of ground texture can provide the LP cue. However, to see the ground surface, the camera must be positioned above the ground, which inevitably introduces the Height-in-Plane (HP) cue. Thus we control Linear Perspective (LP) cue by manipulating the ground texture (\cref{fig:common-pic-cues}). For most cue combinations (except HP/FO and HP/SA), adding the LP cue improves depth ordering accuracies. The relative depth ordering results are summarized in \cref{tab:lp_cue}.

\begin{table}[h]
  \caption{Depth ordering accuracy for images w/ and w/o the Linear Perspective (LP) cue. Introduction of LP improves depth ordering accuracy in majority of the cases. The LP cue is analyzed together with the Height-in-Plane (HP) cue because both of them are the results of raising the camera above the ground. Thus, we report the relative accuracy for LP, in the \textit{Improvement} row.}
  \label{tab:lp_cue}
  \centering
  \setlength{\tabcolsep}{4pt}
  \resizebox{1\textwidth}{!}{%
  \begin{tabular}{@{}lrrrrrrr@{}}
    \toprule
     & HP & HP/FO & HP/LS & HP/OC & HP/RS & HP/SA & HP/TG \\
    \midrule
    Acc. w/o LP  &    0.5579 & 0.5405 & 0.5444 & 0.5734 & 0.5541 & 0.5869 & 0.5560 \\
    Acc. w/ LP     &   0.6081 & 0.5193 & 0.6100 & 0.5753 & 0.6120 & 0.5676 & 0.5849 \\
    \midrule
    Improvement   & 0.0502 & -0.0212 & \textbf{0.0656} & 0.0019 & \textit{0.0579} & -0.0193 & 0.0290 \\
    \bottomrule
  \end{tabular}
  }
\end{table}

\section{Kubric Objects and Environments}

We select 37 objects with different shape complexities from Kubric assets, as listed in \cref{tab:kb_objs}.

\begin{table}[t]
  \caption{37 selected Kubric objects for our rendered images.}
  \label{tab:kb_objs}
    \centering
    \setlength{\tabcolsep}{4pt}
    \resizebox{0.7\textwidth}{!}{%
    \begin{tabular}{ll}
    \toprule
    Asset Source & Object Name \\
    \midrule
    kubasic & cube \\
    kubasic & sphere \\
    kubasic & sponge \\
    kubasic & torus\_knot \\
    kubasic & suzanne \\
    kubasic & spot \\
    gso & Threshold\_Porcelain\_Teapot\_White \\
    gso & W\_Lou\_z0dkC78niiZ \\
    gso & Ecoforms\_Plant\_Container\_Quadra\_Sand\_QP6 \\
    gso & Dog \\
    gso & Nickelodeon\_Teenage\_Mutant\_Ninja\_Turtles\_Michelangelo \\
    gso & Retail\_Leadership\_Summit\_tQFCizMt6g0 \\
    gso & Digital\_Camo\_Double\_Decker\_Lunch\_Bag \\
    gso & Curver\_Storage\_Bin\_Black\_Small \\
    gso & Threshold\_Porcelain\_Pitcher\_White \\
    gso & HyperX\_Cloud\_II\_Headset\_Red \\
    gso & Great\_Dinos\_Triceratops\_Toy \\
    gso & Travel\_Mate\_P\_series\_Notebook \\
    gso & Olive\_Kids\_Birdie\_Sidekick\_Backpack \\
    gso & Mens\_Bahama\_in\_Black\_b4ADzYywRHl \\
    gso & Threshold\_Porcelain\_Coffee\_Mug\_All\_Over\_Bead\_White \\
    gso & Nintendo\_Mario\_Action\_Figure \\
    gso & BABY\_CAR \\
    gso & BIRD\_RATTLE \\
    gso & Black\_Decker\_Stainless\_Steel\_Toaster\_4\_Slice \\
    gso & TriStar\_Products\_PPC\_Power\_Pressure\_Cooker\_XL\_in\_Black \\
    gso & Pennington\_Electric\_Pot\_Cabana\_4 \\
    gso & Crosley\_Alarm\_Clock\_Vintage\_Metal \\
    gso & Threshold\_Bead\_Cereal\_Bowl\_White \\
    gso & Ortho\_Forward\_Facing\_CkAW6rL25xH \\
    gso & Toys\_R\_Us\_Treat\_Dispenser\_Smart\_Puzzle\_Foobler \\
    gso & CoQ10\_BjTLbuRVt1t \\
    gso & Granimals\_20\_Wooden\_ABC\_Blocks\_Wagon\_85VdSftGsLi \\
    gso & ASICS\_GELDirt\_Dog\_4\_SunFlameBlack \\
    gso & Air\_Hogs\_Wind\_Flyers\_Set\_Airplane\_Red \\
    gso & Dino\_5 \\
    gso & Paint\_Maker \\
    \bottomrule
    \end{tabular}
    }
\end{table}

For the Familiar Size (FS) cue, we pick a set of 5 pairs of objects. The two objects in each pair have similar shapes but different sizes. The 5 pairs of objects (with larger ones followed by smaller ones) are:
\begin{itemize}[topsep=0.2em] 
\itemsep0.2em
\tiny \item  Organic\_Whey\_Protein\_Unflavored, CoQ10\_BjTLbuRVt1t
\item \tiny Travel\_Mate\_P\_series\_Notebook, BlackBlack\_Nintendo\_3DSXL
\item \tiny Threshold\_Porcelain\_Pitcher\_White, Threshold\_Porcelain\_Coffee\_Mug\_All\_Over\_Bead\_White
\item \tiny Remington\_TStudio\_Hair\_Dryer, Razer\_Abyssus\_Ambidextrous\_Gaming\_Mouse
\item \tiny TriStar\_Products\_PPC\_Power\_Pressure\_Cooker\_XL\_in\_Black, Threshold\_Porcelain\_Teapot\_White
\end{itemize}

The information of 12 simulated indoor and outdoor environments is presented in \cref{tab:kb_envs}. All of them are rendered with HDRI images, which provide ground, background, and realistic lighting.

\begin{table}[h!]
  \caption{12 selected indoor and outdoor environments, rendered with HDRI images.}
  \label{tab:kb_envs}
  \centering
  \setlength{\tabcolsep}{4pt}
  \begin{tabular}{@{}ll@{}}
    \toprule
     Name & Indoor/outdoor \\
    \midrule
    mud\_road & outdoor \\
    umhlanga\_sunrise & outdoor \\
    aristea\_wreck & outdoor \\
    lenong\_3 & outdoor \\
    evening\_road\_01 & outdoor \\
    waterbuck\_trail  & outdoor \\
    abandoned\_hall\_01 & indoor \\
    aerodynamics\_workshop & indoor \\
    castle\_zavelstein\_cellar & indoor \\
    dresden\_station\_night & indoor \\
    parking\_garage & indoor \\
    royal\_esplanade & indoor \\
    \bottomrule
  \end{tabular}
\end{table}

\pagebreak

\section{Cropping Real-World Images for Depth Ordering}

\begin{figure}[ht]
  \centering
  \begin{subfigure}{0.42\linewidth}
  	\includegraphics[width=\linewidth]{img/img_0692-augmented.jpg}
        \caption{Original}
        \label{fig:si-amc-a}
  \end{subfigure}
  \begin{subfigure}{0.245\linewidth}
  	\includegraphics[width=\linewidth]{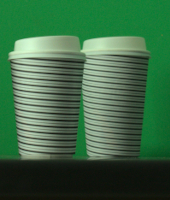}
        \caption{Cropped}
        \label{fig:si-amc-c}
  \end{subfigure}
  \caption{Cropping images.
  (\protect\subref{fig:si-amc-a}) Odd-one-out image before cropping.
  (\protect\subref{fig:si-amc-c}) Cropped images specifically for the depth ordering task.}
  \label{fig:si-aug_crop}
\end{figure}

\cref{fig:si-aug_crop} shows the cropping of real-world images specifically for depth ordering task. Although after cropping the image is no longer odd-one-out, the cropping generates additional images for depth ordering VQA. One such cropped image is used to test both near and far cases, depending which one of the two objects is referred to.

\section{Prompt Variations, Tags, and Examples}

\cref{tab:qvar_example} provides more details on our regular prompt\footnote{These prompts are regular comparing to the prompts used in the in-context learning an chain-of-thought experiments, as presented in the main paper.} formatting.
As shown in the first column of the table, there are 5 types of question variations, \ie clarity, depth, formality, option, 
and query.
The referring clarity variation was described in the main text.
The depth variation concerns how the depth order relation is described, using 3 pairs of antonyms as in the tag names (e.g. \texttt{closer-far}).
The formality variation is about the level of formality of the question, ranging from \texttt{casual} to \texttt{formal}.
The option and query variations are self-explanatory with the examples in \cref{tab:qvar_example}.

\begin{table*}[h!]
  \caption{Question variations by tags. Corresponding variations are shown with \textbf{bold text} in the full prompts.}
  \label{tab:qvar_example}
  \centering
  \setlength{\tabcolsep}{3pt}
  \resizebox{\textwidth}{!}{%
  \begin{tabular}{@{}llp{0.96\linewidth}@{}}
    \toprule
    Variation & Tag name & Full prompt \\
    \midrule
    clarity & \texttt{low} & Is \textbf{the most interesting object} closer to or farther away from the viewer than the rest of the objects? A. Farther. B. Closer. Answer A or B. \\
    clarity & \texttt{med} & Relative to the camera, is \textbf{the object positioned at a different distance} closer or farther away than the remaining objects? A. Closer. B. Farther. Answer A or B. \\
    clarity & \texttt{high} & Is \textbf{the center object} at the rear of the other objects? A. Yes. B. No. Answer A or B. \\
    clarity & \texttt{highest} & Is \textbf{the object marked with a red circle} at the rear of the objects marked with a blue square? A. Yes. B. No. Answer A or B. \\
    \midrule
    depth & \texttt{before-after} & Along the line of sight, is the object marked with a red circle positioned \textbf{before or after} the objects marked with a blue square? A. After. B. Before. Answer A or B. \\
    depth & \texttt{close-far} & Is the object marked with a red circle \textbf{closer to or farther away} from the viewer than the objects marked with a blue square? A. Closer. B. Farther. Answer A or B. \\
    depth & \texttt{front-rear} & Is the special object breaking the depth pattern \textbf{at the rear of} the other similar objects? A. Yes. B. No. Answer A or B. \\
    \midrule
    formality & \texttt{casual} & Does the object in the middle \textbf{feel} farther or nearer than the other similar objects? A. Farther. B. Nearer. Answer A or B. \\
    formality & \texttt{regular} & Is the object in the middle behind or in front of the remaining objects? A. In front of. B. Behind. Answer A or B. \\
    formality & \texttt{formal} & \textbf{Along the line of sight}, is the object in the middle positioned before or after the remaining objects? A. Before. B. After. Answer A or B. \\
    \midrule
    option & \texttt{normal} & Is the object marked with a red circle positioned farther from or closer to the observer than the objects marked with a blue square? A. Closer. B. Farther. Answer A or B. \\
    option & \texttt{reversed} & Is the center object closer to or farther away from the viewer than the other objects? \textbf{A. Farther. B. Closer.} Answer A or B. \\
    \midrule
    query & \texttt{multi-choice} & Relative to the camera, is the center object closer or farther away than the other objects? A. Closer. B. Farther. Answer A or B. \\
    query & \texttt{yes-no} & Is the object marked with a red circle at the rear of the objects marked with a blue square? \textbf{A. Yes. B. No.} Answer A or B. \\
    \bottomrule
  \end{tabular}
  }
\end{table*}

\begin{figure}[h!]
  \centering
  \includegraphics[width=1\linewidth]{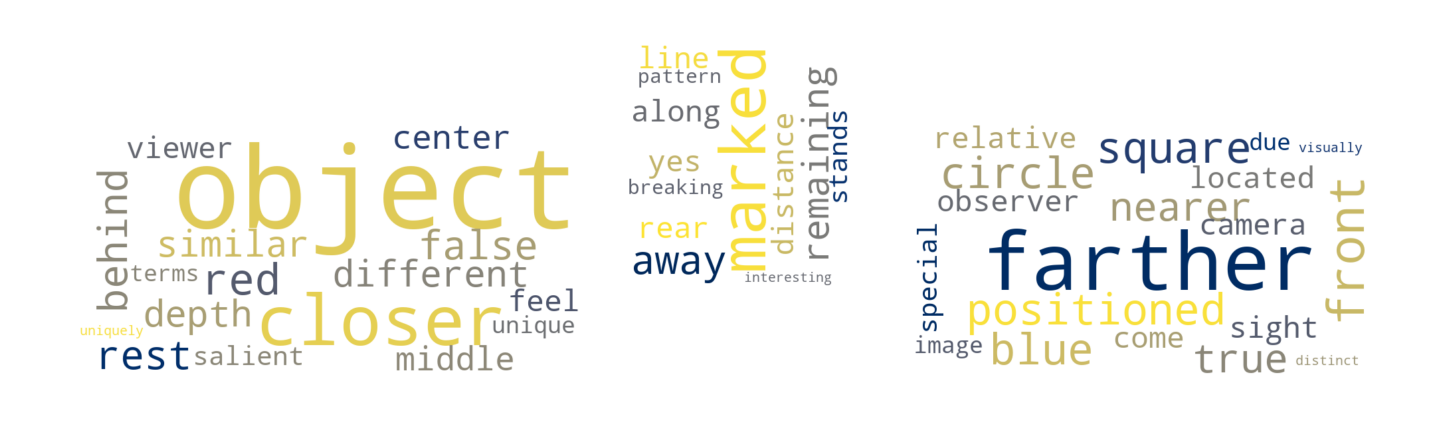}
   \vspace{1em}
   \caption{Word cloud generated from the collection of regular prompts, without ICL and CoT prompting.}
   \label{fig:wcloud}
\end{figure}

While there are 14 unique tags, the combinations of them form 42 unique groups ($\Omega_{L}$ in the main paper). \cref{fig:wcloud} shows a word cloud generated from the regular prompts (w/o ICL and CoT).

Specially for Kosmos2, we have to use \textit{``Answer:''} instead of \textit{``Answer A or B.''} as the \texttt{instruction}, otherwise the model is unable to return meaningful responses for about 70\% of the questions.

\pagebreak

\section{VLM Versions}
\cref{tab:vlms} summarizes the Hugging Face model versions for the evaluated models in the paper, except GPT4.1-mini and Gemini2.5 which are accessed via OpenAI and Google API, respectively.

\begin{table}[h!]
  \caption{Evaluated models and their detailed version information.}
  \label{tab:vlms}
  \centering
  \resizebox{0.6\textwidth}{!}{%
  \begin{tabular}{@{}ll@{}}
    \toprule
    Name & Full version \\
    \midrule
    BLIP2          & \texttt{blip2-flan-t5-xl} \\
    DeepSeek-VL    & \texttt{DeepSeek-VL-7B-chat} \\
    InternVL2.5    & \texttt{InternVL2\_5-4B} \\
    Kosmos2        & \texttt{kosmos-2-patch14-224} \\
    LLaVA1.5       & \texttt{llava-v1.5-7b} \\
    PaliGemma2      & \texttt{paligemma2-3b-mix-224} \\
    Phi3.5         & \texttt{Phi-3.5-vision-instruct} \\
    Qwen2-VL       & \texttt{\small Qwen2-VL-7B-Instruct-GPTQ-Int4}\\
    VILA1.5         & \texttt{VILA1.5-3b} \\
    Cambrian       & \texttt{cambrian-phi3-3b} \\
    GPT4.1-mini     & \texttt{GPT4.1-mini-2025-04-14} \\
    Gemini2.5     & \texttt{gemini-2.5-flash-lite} (July 22, 2025) \\
    \midrule
    DepthAnyV2     & \texttt{Depth-Anything-V2-Small-hf} \\
    \bottomrule
  \end{tabular}
  }
\end{table}

\section{Additional Experiment Results}
This section reports more results for each of the evaluated VLMs. The results are similarly grouped into vision and language dimensions.
In tables that report numeric values (accuracy or SDGM) the \textbf{best} and \textit{second best} results are in bold and italic, respectively.

\subsection{Vision Dimension}

\begin{figure}[h!]
  \centering
  \includegraphics[width=0.9\linewidth]{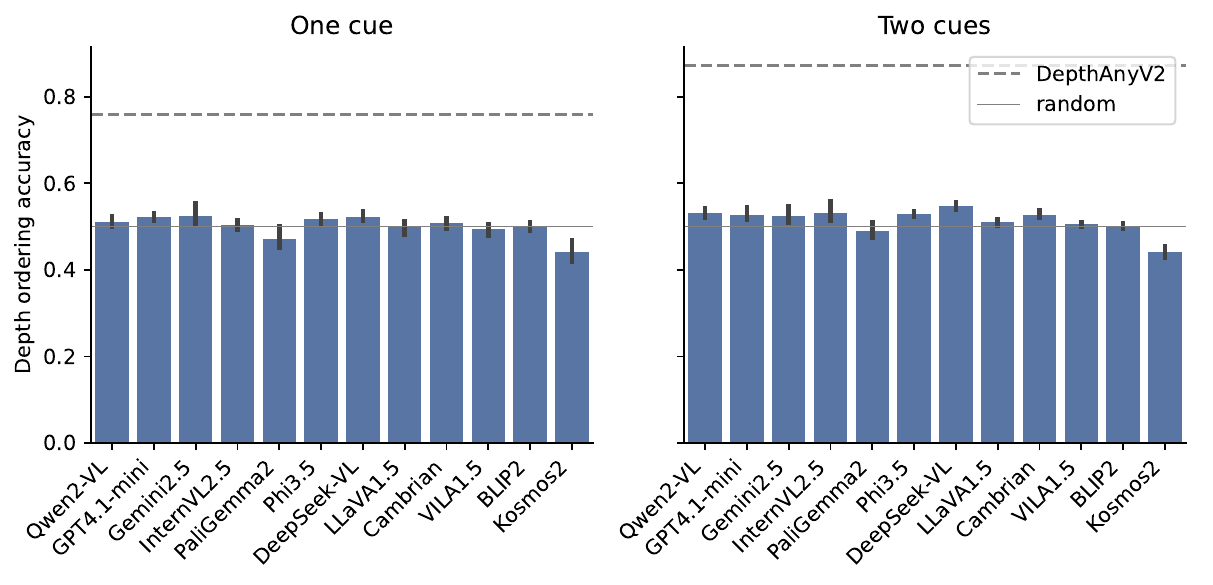}
   \caption{Mean depth ordering accuracy by models and number of cues (aggregated across the 14 environments). Error bars denote $95\%$ CIs. Most VLMs perform at chance level.
   }
   \label{fig:cue_vlms}
\end{figure}

\cref{fig:cue_vlms} shows the performance gap between evaluated VLMs and the DepthAnythingV2 baseline, for single-cue cases and two-cue interactions. 

We report the effects of in-context learning (ICL) and chain-of-thought (CoT) prompting in \cref{tab:vlms_icl} and \cref{tab:vlms_icl_cue}, for each evaluated VLM and each individual cue, respectively. Only the two commercial VLMs take advantage of the ICL and CoT prompting, with the GPT4.1-mini ($\delta=0.0674$) having the largest improvement of depth ordering accuracy.
In terms of individual cues, Relative Size (RS) benefits the most from the ICL and CoT prompting.

\begin{table}[h]
  \caption{The effect of ICL and CoT prompting for 5 of the evaluated VLMs. Only the two commercial VLMs take advantage of the ICL and CoT prompting.}
  \label{tab:vlms_icl}
  \centering
  \setlength{\tabcolsep}{4pt}
  \begin{tabular}{@{}lrr|r@{}}
    \toprule
    VLM & Regular & ICL \& CoT & Improvement \\
    \midrule
    DeepSeek-VL & 0.5264 & 0.5040 & -0.0223 \\
    Phi3.5 & 0.5136 & 0.5072 & -0.0064 \\
    Qwen2-VL & 0.5248 & 0.5128 & -0.0120 \\
    GPT4.1-mini & 0.5371 & 0.6045 & \textbf{0.0674} \\
    Gemini2.5 & 0.4728 & 0.5124 & \textit{0.0396} \\
    \bottomrule
  \end{tabular}
\end{table}
\begin{table}[h!]
  \caption{The effect of ICL and CoT prompting for 8 individual cues. The Relative Size (RS) cue benefits the most from the ICL and CoT prompting.}
  \label{tab:vlms_icl_cue}
  \centering
  \setlength{\tabcolsep}{4pt}
  \begin{tabular}{@{}lrrrrrrrr@{}}
    \toprule
     & HP & OC & RS & TG & SA & LS & FO & FS \\
    \midrule
    Regular & 0.5579 & 0.4910 & 0.5522 & 0.4916 & 0.4872 & 0.4902 & 0.4902 & 0.5379 \\
    ICL \& CoT & 0.5586 & 0.4940 & 0.6214 & 0.4883 & 0.4874 & 0.4940 & 0.4993 & 0.5021 \\
    \midrule
    Improvement & 0.0007 & 0.0030 & \textbf{0.0692} & -0.0034 & 0.0002 & 0.0038 & \textit{0.0091} & -0.0359 \\
    \bottomrule
  \end{tabular}
\end{table}

\cref{tab:vlms_v} provides more details of depth ordering accuracy with respect to cues. When reporting accuracy for individual pictorial cues on the left side of the table, we take the mean of all cue strengths. For the right side of the table dedicated to comparing number of cues, we only use regular cue strength because 2-cue images only have regular strength. The last column presents overall accuracies that were reported in Fig. 4 of the main paper.

\begin{table*}[h!]
  \centering
  \caption{
    Depth ordering accuracy for various pictorial depth cues.
    \textit{Synthetic (all cue strengths)}: mean accuracy across all cue strengths for each individual depth cue (HP: Height-in-Plane, OC: Occlusion, RS: Relative Size, TG: Texture Gradient, SA: Saturation, LS: Light-and-Shadow, FO: Focusness, FS: Familiar Size); \textit{Synthetic}: accuracy averaged over all images with the same number of cues; \textit{Real-world}: accuracy averaged over all images with the same number of cues.
  }
  \vspace{-1em}
  \label{tab:vlms_v}
  \setlength{\tabcolsep}{4pt}
  \resizebox{\textwidth}{!}{%
  \begin{tabular}{@{}lrrrrrrrr|rrr|rrr|r@{}}
    \toprule
    & \multicolumn{8}{c}{Synthetic (all cue strengths)} & \multicolumn{3}{c}{Synthetic} & \multicolumn{3}{c}{Real-world} \\
    \cmidrule{2-9}\cmidrule{10-12} \cmidrule{13-15}
    VLM & HP & OC & RS & TG & SA & LS & FO & FS & 0 cue & 1 cue & 2 cues & 0 cue & 1 cue & 2 cues & Overall \\
    \midrule
    BLIP2 & 0.4979 & 0.4912 & 0.4875 & 0.4865 & 0.4984 & 0.5031 & 0.5078 & 0.5000 & 0.4906 & 0.4984 & 0.5004 & 0.4000 & 0.5294 & 0.5000 & 0.4905 \\
    DeepSeek-VL & 0.5208 & 0.5042 & \textit{0.6237} & 0.5026 & 0.4865 & 0.5099 & 0.4992 & 0.5115 & 0.5010 & \textbf{0.5243} & \textbf{0.5508} & 0.5000 & 0.5000 & 0.5000 & 0.5308 \\
    InternVL2.5 & \textit{0.5863} & 0.5036 & 0.5177 & 0.4912 & 0.4891 & 0.4953 & 0.5166 & 0.5308 & 0.5073 & 0.5009 & 0.5184 & 0.4500 & 0.5294 & \textbf{0.7143} & 0.5438 \\
    Kosmos2 & 0.4423 & 0.4226 & 0.4356 & 0.4215 & 0.4267 & 0.4236 & 0.4621 & 0.4385 & 0.4127 & 0.4299 & 0.4355 & 0.7000 & 0.5882 & 0.5000 & 0.4740 \\
    LLaVA1.5 & 0.5052 & 0.4953 & 0.5120 & 0.4969 & \textit{0.5062} & 0.4943 & 0.5068 & 0.4923 & 0.5042 & 0.5061 & 0.5088 & 0.6000 & 0.4118 & 0.5000 & 0.5275 \\
    PaliGemma2 & 0.4974 & 0.4631 & 0.5172 & 0.4574 & 0.4470 & 0.4449 & 0.4501 & 0.4692 & 0.4376 & 0.4589 & 0.4802 & 0.5000 & \textit{0.6471} & \textit{0.6905} & 0.5386 \\
    Phi3.5 & 0.5231 & \textbf{0.5091} & 0.5330 & 0.5026 & 0.4919 & \textbf{0.5151} & 0.5039 & \textbf{0.5692} & 0.5010 & 0.5172 & \textit{0.5280} & 0.6500 & 0.5294 & 0.6429 & 0.5383 \\
    Qwen2-VL & 0.5585 & 0.4883 & 0.5426 & \textit{0.5070} & 0.5021 & 0.4862 & 0.5000 & 0.5269 & 0.5042 & 0.5068 & 0.5279 & 0.3500 & 0.5588 & 0.5952 & \textbf{0.5577} \\
    VILA1.5 & 0.5000 & \textit{0.5062} & 0.5166 & 0.4948 & \textbf{0.5078} & 0.5021 & 0.5057 & 0.4538 & 0.5218 & 0.4975 & 0.5059 & 0.4500 & 0.4412 & 0.4048 & 0.4960 \\
    Cambrian & 0.5203 & 0.4922 & 0.5920 & 0.4990 & \textit{0.5062} & \textit{0.5135} & 0.4844 & 0.5308 & 0.4834 & 0.5089 & 0.5243 & 0.4500 & 0.5000 & 0.5476 & 0.5193 \\
    GPT4.1-mini & \textbf{0.7230} & 0.4909 & \textbf{0.6279} & 0.4787 & 0.4662 & 0.4875 & 0.4797 & 0.4846 & 0.5052 & \textit{0.5189} & 0.5203 & 0.4500 & 0.5588 & 0.6190 & \textit{0.5546} \\
    Gemini2.5 & 0.5639 & \textit{0.5062} & 0.5210 & \textbf{0.5122} & 0.5008 & 0.4873 & 0.5068 & \textit{0.5346} & 0.4865 & 0.5138 & 0.5126 & 0.3500 & \textbf{0.6765} & \textbf{0.7143} & 0.5502 \\
    \midrule
    VLM mean & \textbf{0.5366} & 0.4894 & \textit{0.5356} & 0.4875 & 0.4857 & 0.4886 & 0.4936 & 0.5035 & 0.4880 & 0.4985 & 0.5094 & 0.4875 & 0.5392 & 0.5774 & 0.5268 \\
    DepthAnyV2 & \textbf{0.9940} & 0.4695 & \textit{0.9084} & 0.5433 & 0.5848 & 0.5120 & 0.5566 & 0.5902 & 0.5177 & 0.6456 & 0.7446 & 0.6667 & 0.9596 & 0.9792 & 0.8475 \\
    \bottomrule
  \end{tabular}
  }
\end{table*}

\begin{figure*}[h!]
  \centering
  \includegraphics[width=\linewidth]{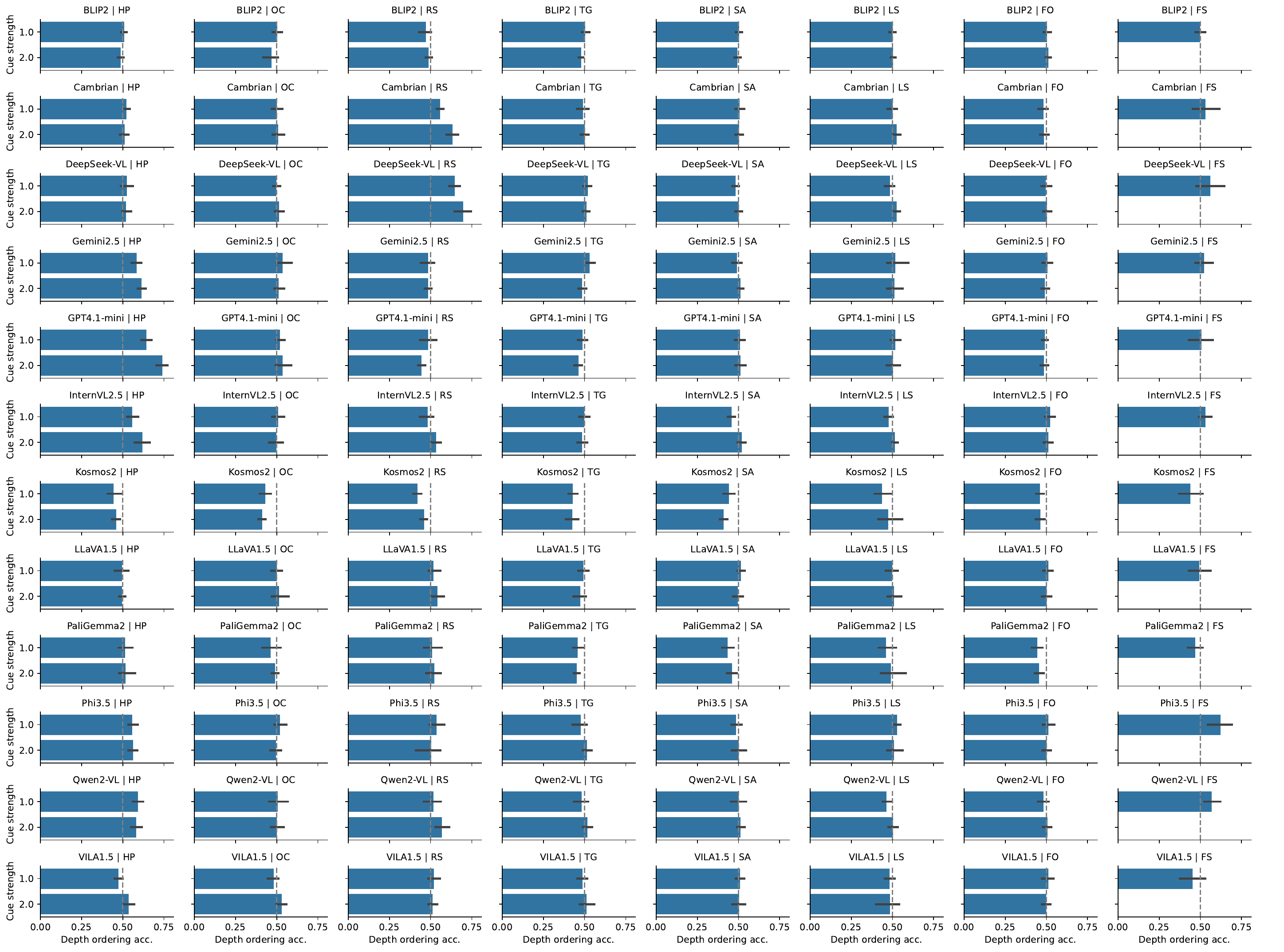}

   \caption{Comparison of depth ordering accuracies as cue strengths double, for each cue and each VLM. Grey dashed lines are chance level (0.5). The double cue strength is not applicable to the Familiar Size (FS) cue.
   }
   \label{fig:cue_strength_vlm}
\end{figure*}

\cref{fig:cue_strength_vlm} plots VLM-level results showing how depth ordering accuracy changes as the strength of each cue doubles. Accuracies for regular and doubled cue strengths are plotted together for easier comparison. Each row in the figure contains plots for a single VLM with columns corresponding to a different depth cue. This figure illustrates no clear trend that cue strength improves accuracy.

\pagebreak
\subsection{Language Dimension}

\cref{tab:var_qs} reports modified SDGMs (\cref{sec:SDGM-spec-qvar}) for 5 types of question variations, showing VLMs' sensitivity to different question formats. The most consistent model is InternVL2.5 ($\sigma=0.1574$). While BLIP2 is the second best ($\sigma=0.1639$), its visual performance is below the chance level (See \cref{tab:vlms_v}), making language consistency less valuable. Specifically, BLIP2 almost always responded with `near' regardless of language and visual cue variations.

\begin{table}[h!]
  \caption{Response inconsistency influenced by question variations. 
  First 5 columns represent modified SDGMs for specific types of question variations (described earlier in \protect\cref{tab:qvar_example}). The last column contains SDGMs across all 42 question variations. Lower SDGM is better.
  }
  \label{tab:var_qs}
  \centering
  \setlength{\tabcolsep}{4pt}
  \begin{tabular}{@{}lrrrrr|r@{}}
    \toprule
    VLM & Clarity & Depth & Formality & Query & Option & $\sigma_{\Omega_L}$\\
    \midrule
    BLIP2         & 0.0779 & \textbf{0.0448} & \textbf{0.0324} & 0.2194 & \textbf{0.0453} & \textit{0.1639} \\
    DeepSeek-VL    & 0.1168 & 0.1679 & 0.1265 & 0.2785 & 0.1005 & 0.3029 \\
    InternVL2.5       & 0.0762 & 0.1415 & 0.0820 & \textbf{0.0600} & \textit{0.0561} & \textbf{0.1574} \\
    Kosmos2         & 0.0662 & 0.3491 & 0.2820 & 0.3433 & 0.1146 & 0.3735 \\
    LLaVA1.5        & \textbf{0.0354} & 0.3523 & 0.1157 & 0.2333 & 0.3190 & 0.3974 \\
    PaliGemma2         & 0.1319 & 0.3396 & 0.1500 & 0.1668 & 0.2520 & 0.3150 \\
    Phi3.5         & 0.1097 & 0.1584 & 0.1266 & 0.1143 & 0.0753 & 0.3113 \\
    Qwen2-VL        & 0.0543 & 0.2002 & 0.0980 & 0.1921 & 0.3719 & 0.3323 \\
    VILA1.5         & \textit{0.0441} & 0.2688 & \textit{0.0492} & 0.1960 & 0.5608 & 0.4282 \\
    Cambrian       & 0.2312 & 0.3089 & 0.2036 & 0.2636 & 0.1028 & 0.3064 \\
    GPT4.1-mini     & 0.0754 & \textit{0.1009} & 0.0930 & \textit{0.0811} & 0.1363 & 0.2719 \\
    Gemini2.5       & 0.0973 & 0.1667 & 0.0626 & 0.1690 & 0.2366 & 0.3039 \\
    \midrule
    VLM mean & \textbf{0.0930} & 0.2166 & 0.1185 & 0.1931 & 0.1976 & 0.3054 \\
    \bottomrule
  \end{tabular}
\end{table}

\begin{table}[h!]
  \caption{Influence of yes-no questions on positive answers.
  Numbers in the two middle columns are target-far accuracy, $\mathrm{Acc}_F$, for multi-choice and yes-no questions, respectively. The difference is defined as \texttt{yes-no} minus \texttt{multi-choice}.
  A positive Diff means the model prefers answering ``yes'' to yes-no questions, and vice versa.
  Only DeepSeek-VL prefers answering ``no''.}
  \label{tab:vlm_yn_delta}
  \centering
  \setlength{\tabcolsep}{4pt}
    \resizebox{0.75\textwidth}{!}{%
  \begin{tabular}{@{}lrr|r@{}}
    \toprule
    VLM & $\mathrm{Acc}_F$ @ \texttt{multi-choice} & @ \texttt{yes-no} & Diff \\
    \midrule
    BLIP2 & 0.0398 & 0.6804 & 0.6406 \\
    DeepSeek-VL & 0.5824 & 0.3251 & -0.2573 \\
    InternVL2.5 & 0.2993 & 0.5373 & 0.2380 \\
    Kosmos2 & 0.2381 & 0.4959 & 0.2578 \\
    LLaVA1.5 & 0.3245 & 0.8489 & 0.5244 \\
    PaliGemma2 & 0.4749 & 0.4909 & 0.0160 \\
    Phi3.5 & 0.3468 & 0.8995 & 0.5527 \\
    Qwen2-VL & 0.3753 & 0.8974 & 0.5221 \\
    VILA1.5 & 0.4523 & 0.5082 & 0.0559 \\
    Cambrian & 0.5607 & 0.5650 & 0.0042 \\
    GPT4.1-mini & 0.3753 & 0.4057 & 0.0304 \\
    Gemini2.5 & 0.2962 & 0.3705 & 0.0743 \\
    \midrule
    VLM mean & 0.3638 & 0.5854 & 0.2216 \\
    \bottomrule
  \end{tabular}
  }
\end{table}

\cref{tab:vlm_yn_delta} shows the influences of yes-no questions on the positive answers. Numbers in the two middle columns are target-far accuracy, $\mathrm{Acc}_F$, defined on subsets of images where the targets are farther away than the distractors.
We frame the yes-no questions as ``Is the target behind distractors?'', and the positive answer means ``far''. Therefore when a model prefers answering ``yes'', the $\mathrm{Acc}_F$ of \texttt{yes-no} should increase comparing to \texttt{multi-choice}. As shown in the difference column, most VLMs prefer ``yes'', except DeepSeek-VL.
Note that the target-far accuracies themselves are less meaningful than the difference as VLMs have different degrees of near-far biases. 

\begin{table*}[ht!]
  \caption{Depth ordering accuracy with respect to referring clarity. The VLM mean improvements due to increased clarity are marginal. Kosmos2's referring by bounding box does not boost accuracy either.
  }
  \label{tab:vlms_l_cl}
  \centering
  \setlength{\tabcolsep}{4pt}
  \begin{tabular}{@{}lrrrrr|r@{}}
    \toprule
     & \multicolumn{5}{c}{Synthetic} & \multicolumn{1}{c}{Real-world} \\
    \cmidrule{2-6}\cmidrule{7-7}
    VLM & Low & Med & High & Highest & BBox & Highest \\
    \midrule
    BLIP2 & 0.4917 & 0.5032 & 0.4995 & 0.4987 &- & 0.4943 \\
    DeepSeek-VL & 0.4807 & 0.5133 & 0.5219 & \textbf{0.5438} &- & 0.5057 \\
    InternVL2.5 & 0.5016 & 0.5076 & 0.5232 & 0.5168 &- & 0.5805 \\
    Kosmos2 & 0.3878 & 0.3979 & 0.4305 & 0.4347 & 0.4671
 & 0.5172 \\
    LLaVA1.5 & 	0.5005 & 0.4989 & 0.4992 & 0.5063 &- & 0.5115 \\
    PaliGemma2 & 0.4807 & 0.5043 & 0.5108 & 0.4755 &- & 0.5977 \\
    Phi3.5 & 0.4953 & 0.5035 & 0.5138 & \textit{0.5234} &- & 0.5345 \\
    Qwen2-VL & \textbf{0.5134} & \textit{0.5176} & \textit{0.5343} & 0.5225 &- & 0.5805 \\
    VILA1.5 & \textit{0.5035} & 0.5015 & 0.5097 & 0.5052 &- & 0.4655 \\
    Cambrian & 0.4874 & 0.5025 & 0.4733 & 0.5208 &-  & 0.5230 \\
    GPT4.1-mini & 0.4925 & \textbf{0.5347} & \textbf{0.5559} & 0.5207 &- & \textit{0.6034} \\
    Gemini2.5 & 0.4327 & 0.4076 & 0.4979 & 0.5128 &- & \textbf{0.6149} \\
    \midrule
    VLM mean & 0.4807 & 0.4911 & \textit{0.5058} & \textbf{0.5068} &- & 0.5441 \\
    \bottomrule
  \end{tabular}
\end{table*}

\cref{tab:vlms_l_cl} shows depth ordering accuracy regarding referring clarity for synthetic and real-world images. Kosmos2 supports referring by bounding boxes via a special syntax in the prompt, but it does not boost the accuracy.

\pagebreak
\section{Single-Cue Correlation}

\begin{figure}[h!]
  \centering
    \vspace{-1em}
  \includegraphics[width=0.5\linewidth]{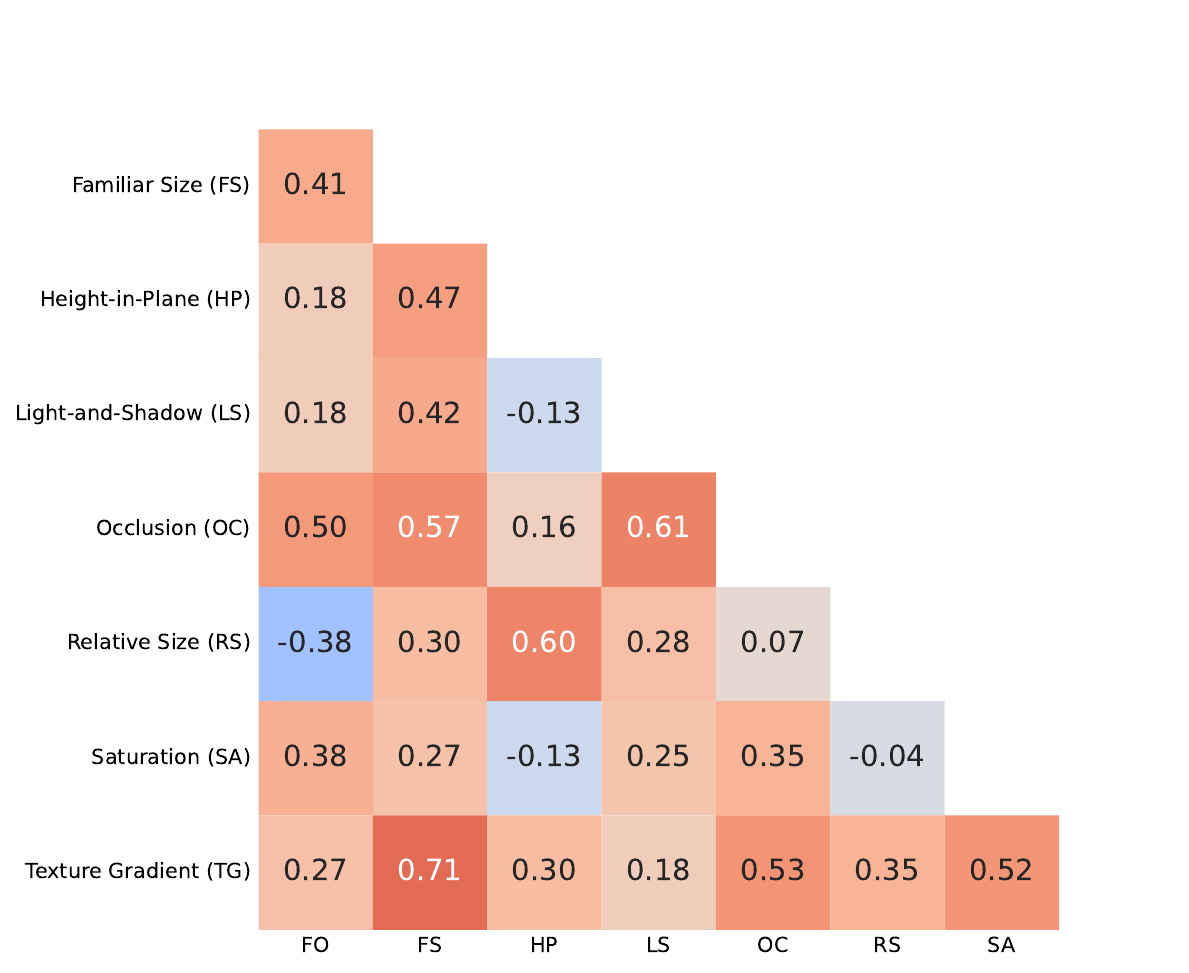}
    \vspace{-1em}
   \caption{A heatmap showing Spearman's rank correlation among single cues.
   The ranks of the depth ordering accuracies of VLMs are used to compute the correlation.
   }
   \label{fig:spearman-cues}
\end{figure}

\cref{fig:spearman-cues} shows Spearman's rank correlations among the 8 single-cue depth ordering accuracies.
The Texture Gradient/Familiar Size (0.71) correlates the most among all pairs, while the Relative Size/Focusness pair (-0.38) has the lowest correlation.
The correlations are consistently lower than those in DepthCues \cite{danierDepthCuesEvaluatingMonocular2025}.

\section{Standard Deviation of Within-Group Means (SDGM) for a Specific Type of Question Variation}
\label{sec:SDGM-spec-qvar}
For context, the last column of \cref{tab:var_qs} reports the SDGM for all 42 question variations formed by the tags (\protect\cref{tab:qvar_example}) from the 5 variation types. To measure the influence of a specific type of question variation (\eg \texttt{yes-no} vs \texttt{multi-choice} query), we modify the SDGM metric. We use the `query' variation (\texttt{yes-no} vs \texttt{multi-choice}) in this section throughout for a more concrete explanation without losing generality.
The original formulation of SDGM is given by $\sigma_{\Omega_\texttt{q}}(\mu) = \sqrt{\frac{1}{||\Omega_\texttt{q}||} \sum_{g \in \Omega_\texttt{q}}(\mu_g - \bar{\mu})^2}$ where $\Omega_\texttt{q} = \Omega_\texttt{query} := \{\mathrm{yes\_no}, \mathrm{multi\_choice}\}$. We define the modified SDGM as the mean of a set of SDGMs
\begin{equation}
  \sigma'_{\Omega_\texttt{q}} =  \frac{1}{||\Omega_{\texttt{q}-}||} \sum_{g \in \Omega{\texttt{q}-}}\sigma_{\Omega_\texttt{q}^g}.
  \label{eq:SDGM-modified}
\end{equation}
Here, $\Omega_{\texttt{q}-}$ denotes the question groups formed by all types of question variations \emph{except} current type $\texttt{q}$, \ie 
\begin{equation*}
\{\mathrm{(low,casual,close\_far,normal)}, \mathrm{(low,casual,close\_far,reversed)}, ...\} 
\end{equation*}
Then, the regular SDGM is computed for each $g \in \Omega_{\texttt{q}-}$. This per-group SDGM is denoted by $\sigma_{\Omega_\texttt{q}^g}$, where the superscripted $g$ distinguishes that the calculation is applied to a group, not the entire data.

The advantage of the modified SDGM is that it captures the influence of the specific type of question variations, \emph{without mixing the influences of the other types}.
This non-mixing property is critical when the other types can cancel the influence of the current type. For example, VILA1.5 mostly responded with the first option regardless of question variations. As shown in the bottom two rows of \cref{tab:qvar_example}, the first option of \texttt{yes-no} (``Is it behind?'') \vs \texttt{multi-choice} (``Is it closer or farther?'') have opposite meanings, \ie \emph{behind} and \emph{closer}, respectively. For the target-near accuracy metric $\mathrm{Acc}_N$, a good SDGM for VILA1.5 should be large to capture the high variation. The original SDGM, $\sigma_{\Omega_\texttt{query}}(\mu) = \sqrt{\frac{1}{2} [(\mu_{\mathrm{yn}} - \bar{\mu})^2+(\mu_{\mathrm{mc}} - \bar{\mu})^2]}$, could have reflected the instability, if \emph{the option question variations did not randomize the option orders} (via \texttt{normal} \vs \texttt{reversed}). The modified SDGM, on the other hand, is able to represent the variation of each finer group $g \in \Omega_{\texttt{q}-}$, \eg $\mathrm{(low,casual,close\_far,normal)}$ within which the option order is not randomized.

We compare the original and modified SDGM implementations in \texttt{pandas} as follows:

{\small
\begin{verbatim}
# original SDGM
(data
    .groupby(["query"])  # G in SDGM
    .depth_ordering_accuracy.mean()  # M
    .std()  # SD
)

# modified SDGM
(data
    .groupby(["clarity", "depth", "formality", "option", "query"])
    .depth_ordering_accuracy.mean()
    .groupby(["clarity", "depth", "formality", "option"])
    .std()
    .mean()
)
\end{verbatim}
}

\end{document}